\documentclass{article} 
\usepackage{iclr2025_conference,times}

\usepackage{amsmath,amsfonts,bm}

\def\eqref#1{equation~\ref{#1}}

\def\1{\bm{1}}

\DeclareMathAlphabet{\mathsfit}{\encodingdefault}{\sfdefault}{m}{sl}
\SetMathAlphabet{\mathsfit}{bold}{\encodingdefault}{\sfdefault}{bx}{n}

\definecolor{citecolor}{HTML}{0071bc}

\usepackage{hyperref}
\usepackage{mathabx}    
\usepackage{xcolor}         
\usepackage{amsmath}
\usepackage{tabularx}
\hypersetup{hidelinks}
\usepackage{wrapfig}

\usepackage{array}
\usepackage{bm}
\usepackage{stfloats}
\usepackage{amsmath}
\usepackage{amssymb}
\usepackage{mathtools}
\usepackage{amsthm}
\usepackage[ruled,vlined]{algorithm2e}  
\usepackage{mathabx}    
\usepackage{mathrsfs,algorithmic}  

\usepackage[utf8]{inputenc} 
\usepackage[T1]{fontenc}    
\usepackage{booktabs}       
\usepackage{amsfonts}       
\usepackage{nicefrac}       
\usepackage{microtype}      
\usepackage{xcolor}         
\usepackage{bbm, dsfont}
\usepackage{wrapfig}
\usepackage{wrapfig}
\usepackage{graphicx}
\usepackage{multirow}
\usepackage{booktabs}
\usepackage{tikz}
\usepackage{comment}
\usepackage{amsmath,amssymb} 
\usepackage{color}
\usepackage{bm}
\usepackage{makecell}
\usepackage{enumitem}
\usepackage{colortbl}
\definecolor{gray}{gray}{.9}
\usepackage{caption}
\usepackage{subcaption}

\hypersetup{hypertex=true,
    colorlinks=true,
    linkcolor=citecolor,
    anchorcolor=citecolor,
    citecolor=citecolor}

\theoremstyle{plain}

\theoremstyle{definition}

\iclrfinalcopy

\title{Data Pruning by Information Maximization} 

\author{
Haoru Tan \thanks{
Equal Contribution: \texttt{\{tanhr2014, stonewst\}@163.com} \\\  $^\dagger$ Corresponding Author 
} $^{\hspace{1.5mm} 1}$,\,\, Sitong Wu$^{* \ 2}$,\,\, Wei Huang$^{1}$,\,\, Shizhen Zhao$^{1}$,\,\, \textbf{Xiaojuan Qi}$^{\dagger \ 1}$
  \vspace{1em} \\
\,\,$^1$The University of Hong Kong \,\,~~~~~~$^2$The  Chinese University of Hong Kong\vspace{0.1em} \\
  }

\begin{document}

\maketitle

\begin{abstract} 
In this paper, we present InfoMax, a novel data pruning method, also known as coreset selection, designed to maximize the information content of selected samples while minimizing redundancy. By doing so, InfoMax enhances the overall informativeness of the coreset. The information of individual samples is measured by importance scores, which capture their influence or difficulty in model learning. To quantify redundancy, we use pairwise sample similarities, based on the premise that similar samples contribute similarly to the learning process.
We formalize the coreset selection problem as a discrete quadratic programming (DQP) task, with the objective of maximizing the total information content, represented as the sum of individual sample contributions minus the redundancies introduced by similar samples within the coreset.
To ensure practical scalability, we introduce an efficient gradient-based solver, complemented by sparsification techniques applied to the similarity matrix and dataset partitioning strategies. 
This enables InfoMax to seamlessly scale to datasets with millions of samples. 
Extensive experiments demonstrate the superior performance of InfoMax in various data pruning tasks, including image classification, vision-language pre-training, and instruction tuning for large language models. Code is available at \url{https://github.com/hrtan/InfoMax}. 
\end{abstract}

\section{Introduction}
\label{sec:introduction}


\begin{wrapfigure}{r}{0.48\textwidth}
    \centering
    \vspace{-1.35786cm}
    \hspace{-0.01245613518486cm}
    \includegraphics[width=0.48\textwidth]{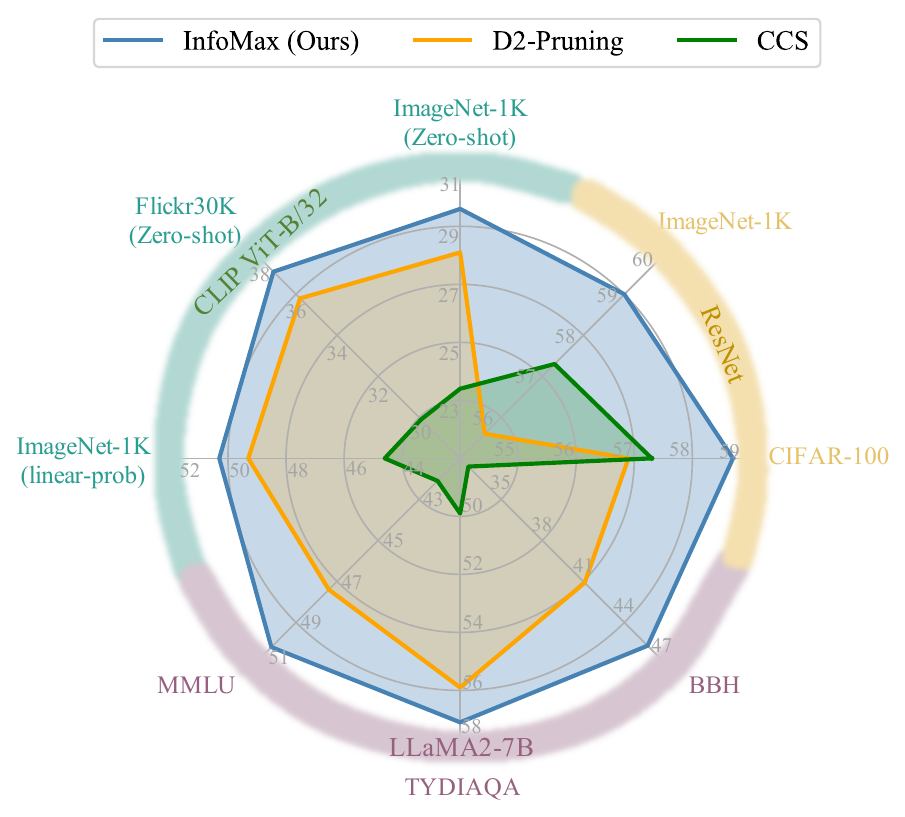}
    \vspace{-0.386cm}
    \caption{\label{fig: intro comparison performance} Summarization of InfoMax's performance in vision-language pre-training and image classification (both at 10\% selection ratio), and instruction fine-tuning for large language models (5\% selection ratio). The results show that InfoMax has demonstrated substantial progress in various scenarios, see Section \ref{sec: experiments} for details.
    \vspace{-0.386cm}
    }
\end{wrapfigure}
Large-scale datasets have been pivotal in the recent breakthroughs in deep learning \citep{GPT3, SAM, CLIP, SD}. However, the growing size of training data substantially increases both training costs and storage demands. 
Moreover, significant redundancies within these datasets highlight the importance of data-pruning methods that remove redundant samples and identify a compact yet informative subset, known as a coreset, to enable more efficient model training and data storage.

\begin{figure*}[tp]
\centering 
\includegraphics[width=1\linewidth]{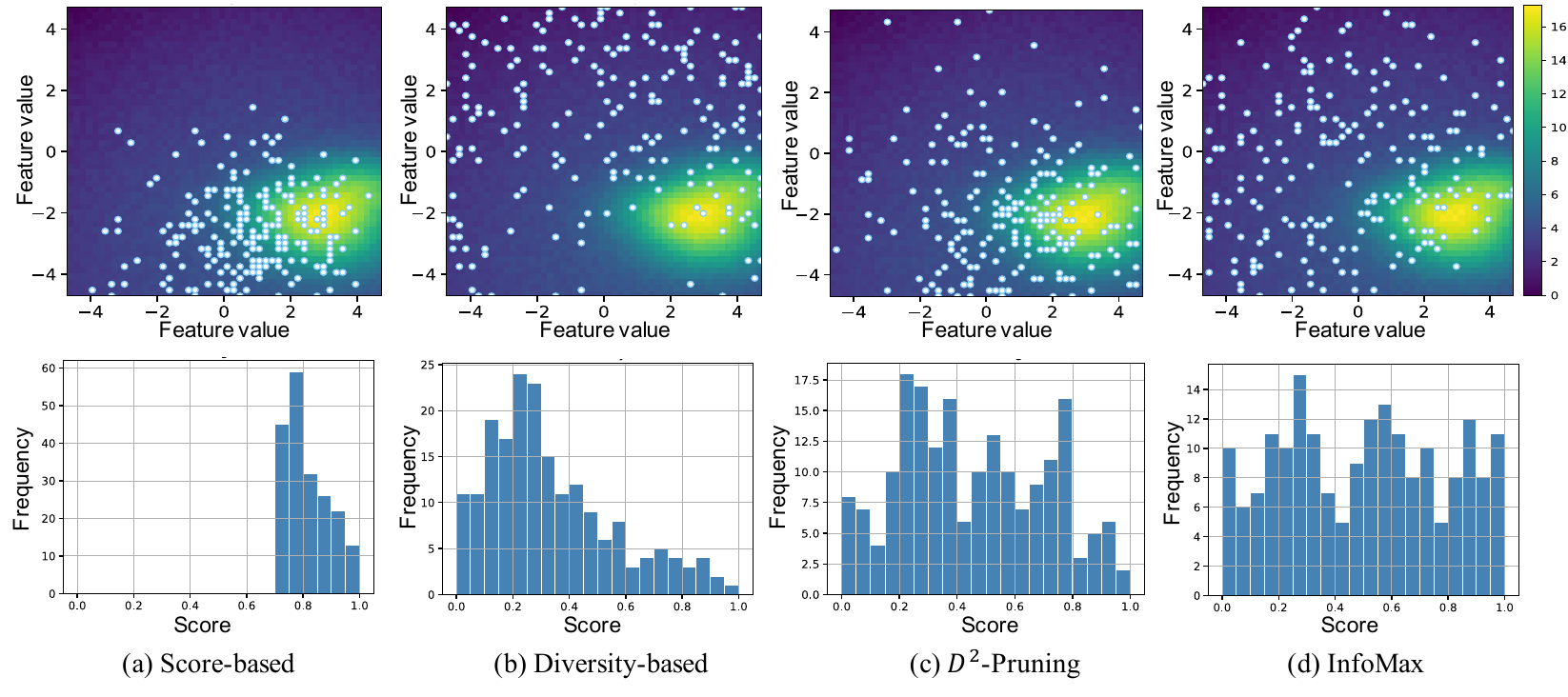}
\vspace{-0.5cm}
\caption{\label{fig: comp_scatter_infomax}
The coreset distributions from InfoMax, and, the score-based method (output margin \citep{margin}, and diversity-based method (k-Median clustering \citep{feldman2011unified,coreset}), hybrid method ($D^2$-Pruning \citep{d2pruning}). 
In the upper figures, the background illustrates the distribution density of the data in the space after PCA dimensionality reduction. Brighter colors indicate a higher density of samples. Simultaneously, we present the locations of the samples selected by different methods using scatters. 
The figures below show the distribution of the scores of the corresponding coreset. This set of experiments is conducted on CIFAR-100 \citep{CIFAR}. The coreset selected by our InfoMax is capable of covering both high-density and low-density regions in terms of spatial uniformity, and also high-information and low-information data in terms of score distribution. 
} 
\vspace{-0.7cm}
\end{figure*}

Research in this field can be broadly divided into two categories: score-based methods \citep{SSP, CLIP} and geometry-based methods \citep{k_center, BADGE}.
Score-based methods focus on developing metrics to evaluate a sample's informativeness, such as prediction uncertainty \citep{margin}, loss value \citep{entropy}, or influence score \citep{xia2024less, tan2023data}. However, as shown in Figure \ref{fig: comp_scatter_infomax}(a), these methods often select samples densely concentrated in regions with the highest scores, leading to redundancies and failing to consider simpler samples with lower scores, which results in biased selections. Recent work \citep{SSP} highlights that even simple samples are important for improving model generalization.
On the other hand, geometry-based methods aim to select a diverse subset of the data, reducing redundancies between samples \citep{k_center}. As illustrated in Figure \ref{fig: comp_scatter_infomax}(b), while these methods prioritize diversity, they often overlook informative samples with high-importance scores, leaving a large number of low-scoring samples in the selection. 

Recently, hybrid approaches have emerged that combine importance scores with diversity to design more effective algorithms \citep{BADGE, d2pruning, CCS,yang24icmlccsimporve}. One of the most notable examples is $D^2$-Pruning \citep{d2pruning}, which models a dataset as a graph. In this framework, node scores represent a sample's informativeness, such as difficulty or importance, while edges capture the similarities between samples. The data pruning process is formulated as an iterative node selection problem, where at each step, nodes with the highest scores are selected, and the scores of neighboring nodes are reduced to account for redundancy. 
However, due to its greedy selection process, the algorithm is prone to getting stuck in suboptimal solutions, making it challenging to maintain a proper balance between importance and diversity. For example, as shown in Figure \ref{fig: comp_scatter_infomax}(c), many samples remain concentrated in high-density areas with low scores, leaving significant portions of the space uncovered. 



\begin{wrapfigure}{r}{0.38\textwidth}
    \centering
    \vspace{-0.486cm}
    \includegraphics[width=0.38\textwidth]{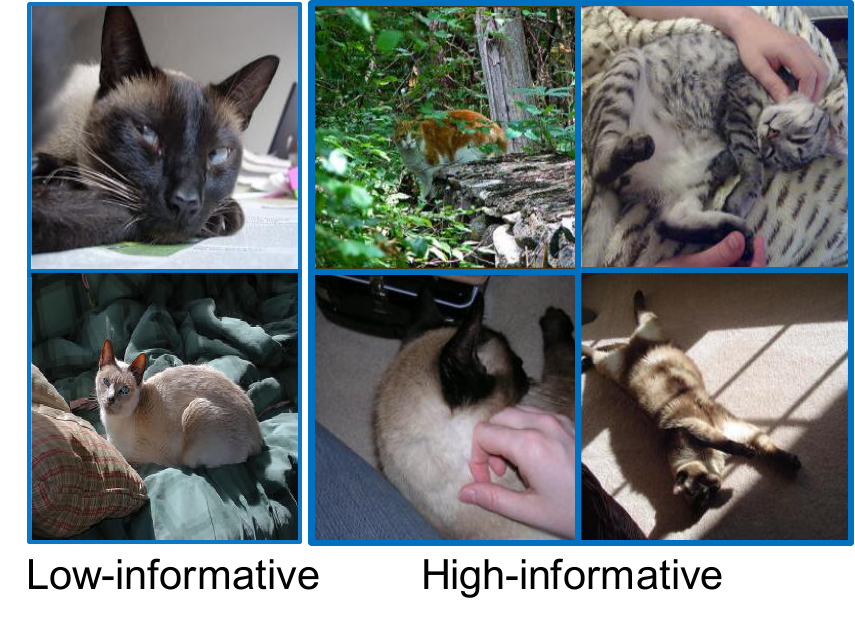}
    \vspace{-0.5186cm}
    \caption{\label{fig: what_information} Cases of samples with low informativeness (left) and high informativeness (right).
    \vspace{-0.486cm}
    }
\end{wrapfigure}
\begin{figure*}[tp] 
\centering
\hspace{-0.2cm}
\includegraphics[width=1\linewidth]{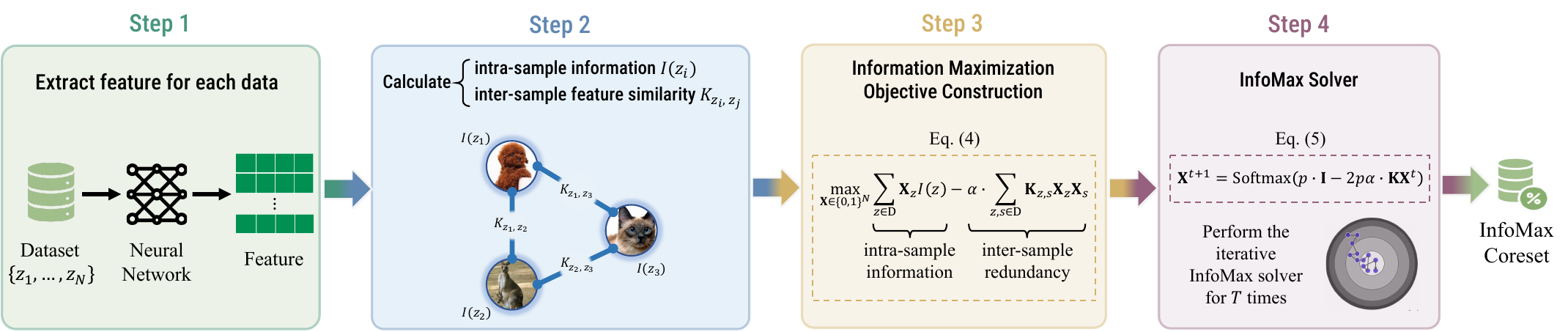}
\caption{\label{fig:pipeline}
The overall pipeline of our InfoMax. In the first two steps, we use a network to extract features to calculate the similarity matrix and the intra-sample information terms. 
Then, in step 3, we construct the quadratic optimization problem for infoMax, see Eq.~(\ref{eq: quadratic}) for details. 
Finally, we perform the iterative InfoMax solver as defined by Eq.~(\ref{eq:update rule}) to obtain the final InfoMax coreset. 
} \vspace{-0.5cm}
\end{figure*}

In this paper, we introduce \emph{InfoMax}, a new and effective approach for data pruning and coreset selection. Our core insight is to find a subset of samples that maximizes overall information by simultaneously considering each sample's information contribution and the information overlap among them. 
First, a sample's informativeness can be explained by its importance or difficulty. For instance, a sample with intricate structures, complex backgrounds, and occlusions provides more information than one with simple patterns see Figure \ref{fig: what_information}. Therefore, we explore score-based metrics that evaluate a sample's difficulty or importance, providing an information assessment. 
Second, the information overlap among samples is evaluated in a pairwise manner and quantified by their pairwise similarities. Samples with greater similarity will have a higher degree of information overlap. 
Using these considerations, we formulate coreset selection as a discrete quadratic programming (DQP) problem with equality constraints that specify the desired number of selected samples, see Eq.~(\ref{eq: quadratic}). The objective function represents the overall information as the sum of each sample's individual information, reduced by redundancies introduced by the presence of similar samples within the coreset.
Furthermore, we propose an efficient and robust gradient-based solver to address scalability. This solver is enhanced by sparsification techniques applied to the similarity matrix and dataset partitioning strategies, enabling practical and scalable computation. For instance, our method processes 12 million data points in just 37 minutes (see Section \ref{sec: time} for details). 
By solving this problem, InfoMax identifies a subset of samples that maximizes overall information, thereby reducing the likelihood of suboptimal solutions often seen with greedy-based methods, see the discussion in Section \ref{sec: How does InfoMax optimize the most informative coreset?}. 
As shown in Figure \ref{fig: comp_scatter_infomax}(b), the coreset generated by our approach strikes a good balance between diversity (i.e., well-distributed across the entire space) and informativeness (i.e., containing a reasonable number of important samples). 
The superior performance across multiple tasks, including image classification, vision-language pretraining, and instruction fine-tuning for large language models, see Section \ref{sec: experiments}, further supports the effectiveness of our new formulation. 
We have summarized the overall pipeline in Figure \ref{fig:pipeline} and Algorithm \ref{alg}. 
In summary, our contributions are:

\begin{itemize}[leftmargin=0.4cm] 
\item  We propose InfoMax, a new coreset algorithm designed to maximize overall information by accounting for each sample's individual contribution while reducing information overlap, with a simultaneous focus on maintaining diversity and importance. 
\item We propose an efficient gradient-based solver enhanced by sparsification techniques and dataset partitioning strategies to make InfoMax scale to large-scale datasets.

\item Extensive experiments show that InfoMax exhibits the best performance and consistently outperforms the state-of-the-art schemes in a series of tasks, including image classification, and vision-language pre-training \citep{CLIP}, large language model supervised fine-tuning \citep{xia2024less} experiments. Notably, it brings about a significant improvement of approximately 5.5\% compared to the previous best methods at a 5\% selection rate in instruction fine-tuning experiments. Additionally, it shows around 2\% performance enhancements on classification tasks at a 10\% selection rate, see Figure \ref{fig: intro comparison performance}. 
\end{itemize}

\section{Preliminaries}
\label{sec:related works} 

In this section, we first present the problem definition for data pruning, followed by a discussion of existing methods, including score-based, diversity-based, and hybrid approaches. 

\subsection{Problem Definition}
Before diving into the literature review of existing methods, we first define the problem of data pruning, also known as coreset selection \citep{k_center, gradient_1, gradient_2}. Let $D = {(z_i)}_{i=1}^N$ represent the training set, drawn independently and identically distributed (i.i.d.) from an underlying data distribution.
The goal of dataset pruning is to identify the most informative subset of the training data that minimizes information loss while maximizing model performance. Formally, the problem can be stated as:
\begin{equation} 
\mathbf{S}^{*} = \arg\max_{\mathbf{S} \subset \mathbf{D}, |\mathbf{S}|=p} I(\mathbf{S}),
\label{eq:true info}  
\end{equation}
where $p$ is the budget of the target coreset and $|\cdot|$ represents the cardinality of a set. $I(\mathbf{S})$ measures the set-level information of the candidate subset $S$. 
There are multiple choices \citep{SSP,Submodularity_1,Submodularity_2} for the instantiation for the set information $I(\mathbf{S})$. For example, 
given the loss function $l(\cdot)$ and the network parameters $\mathbf{w}$, 
previous works \cite{OPT,CCS} often use the overall test loss reduction caused by training the model on $S$ as the information measure $I(\mathbf{S})=\mathbb{E}_{z \sim P} \Big[l (z, \mathbf{w}^0) - l(z, \mathbf{w}_{\mathbf{S}}^*) \Big]$,   Eq.~(\ref{eq:true info}) would be identical as ${\mathbf{S}}^{*} = \underset{\mathbf{S} \subset D, |\mathbf{S}|=p}{\arg\min}~\mathbb{E}_{z \sim P}\Big[ l(z, \mathbf{w}_{\mathbf{S}}^*)  \Big]$, where $\mathbf{w}^0$ is the initial parameter and $\mathbf{w}_{\mathbf{S}}^*$ indicates the optimal network parameter learned on $\mathbf{S}$.  
The key distinction between various data pruning methods lies in how they define $I(\mathbf{S})$, which is detailed below.

\subsection{Score-based Method} 

The score-based method \citep{GraNd_EL2N,tan2023data,forgetting} often selects samples solely based on the score values. They generally model $I(\mathbf{S})$ as: $I(\mathbf{S}) = I(z_1) + I(z_2) + ... + I(z_p)$, 
where $\mathbf{S} = \{z_1, \dots, z_p\} \subset \mathbf{D}$, with each $z$ representing a data sample, and $I(z)$ denoting the information associated with that sample.
The primary task then becomes identifying a suitable score metric to evaluate a sample's individual contribution, $I(z)$, of each sample. Various scores have been proposed to quantify $I(z)$, such as model uncertainty \citep{margin}, loss value \citep{entropy}, and influence score \citep{tan2023data,TracIN}. 

The assumption underlying the score-based formulation is that there is no information overlap between different samples, allowing the individual contributions of samples to be summed to represent the overall contribution. However, this approach fails to account for overlap in the information provided by different samples. For instance, if two identical samples, $z_i$ and $z_j$, both receive high scores, the information gained from adding $z_j$ after selecting $z_i$ would be negligible. As a result, this method cannot ensure that the selected samples offer broad coverage of the data space, leading to a loss of diversity and suboptimal solutions. This limitation is illustrated in Figure \ref{fig: comp_scatter_infomax}(a), they select samples densely concentrated in regions with the highest scores, leading to redundancies and failing to consider simpler samples with lower scores, which results in biased selections.

\subsection{Geometry-based Method and Hybrid Method} 
\paragraph{Geometry-based Method} In contrast to score-based methods, geometry-based methods design   $I(\mathbf{S})$ to consider maximizing the diversity and minimizing the information overlap among samples. 
For many works \citep{lloyd1982least,tan2006cluster,coates2012learning,harpeled2004on,feldman2011unified}, the set-level information $I(\mathbf{S})$ is regarded as: $I(\mathbf{S})~ \propto \sum_{(z_i, z_j) \in \mathbf{S}} -I(z_i; z_j)$,
where $I(z_i, z_j)$ measures the similarity of two samples, indicating information overlap (also the mutual information).  To solve the above problem,  \cite{k_center} applied greedy k-center to choose the coreset with good data coverage. 

This formulation assumes that all individual samples carry equal amounts of information, disregarding the varying significance of each sample. Consequently, it tends to overlook critical samples while retaining a large number of non-informative ones. As illustrated in Figure \ref{fig: comp_scatter_infomax}(b), while these methods prioritize diversity, they often overlook informative samples with high-importance scores, leaving a large number of low-scoring samples in the selection. 

\paragraph{Hybrid Method} More recently, hybrid methods \citep{CCS,BADGE,d2pruning} have been developed that account for both the individual importance and diversity of samples simultaneously.  For instance, CCS \citep{CCS} sets different sampling ratios for samples with different scores to enhance data coverage and balance both easy and hard samples. 
$D^2$-Pruning \citep{d2pruning} proposes to select data on the graph. One of its core steps is the Inverse Message Passing operation. Specifically, it selects the sample with the highest score among all unselected candidates and subsequently reduces the scores of the neighborhood. 
However, the above approaches are primarily based on heuristic designs and are often solved using greedy algorithms. These methods are prone to local optima, as the algorithm lacks a holistic view of the problem at each step, resulting in suboptimal outcomes. In Figure \ref{fig: comp_scatter_infomax}, the selected samples demonstrate inadequate coverage of the entire space, with many concentrated in low-importance regions, hence results in the suboptimal performance in practice, see Figure \ref{fig: intro comparison performance} for details. 

In this paper, we propose a unified formulation with a scalable solver for coreset selection, aimed at maximizing the information of the selected subset by accounting for both the individual contributions of samples and their redundancies. 

\section{Method}

\subsection{InfoMax: Formulations and Solutions} 
\label{sec: forms and solutions}

The objective of InfoMax is to identify a subset of data samples that maximize intra-sample information while minimizing inter-sample redundancies caused by similar samples, thereby achieving an optimal balance between diversity and importance.  For a sample $z$, we represent its information content as $I(z)$. For two samples $z$ and $s$, their information overlap or redundancy is denoted as $I(z,s)$. The total information is measured by summing the individual information of the samples, with deductions for inter-sample redundancy, as outlined below.  
We will discuss in Section \ref{sec: How does InfoMax optimize the most informative coreset?} that under some mild settings, this optimization problem in Eq.~(\ref{eq: quadratic}) is equivalent to solving the original information maximization problem for data pruning as defined in Eq.~(\ref{eq:true info}). 


\paragraph{Problem Formulation for InfoMax} Here, we consider inter-sample redundancy in a pairwise manner using $I(z,s)$ and the matrix containing all pairwise sample similarities is denoted as $\mathbf{K}_{z,s}$. Then, given the dataset $\mathbf{D} = \{z_1, z_2,...,z_N\}$,  we introduce a variable $\mathbf{X}_z \in \{0,1\}$ to represent whether a sample is selected ($\mathbf{X}_z = 1$) or pruned ($\mathbf{X}_z = 0$), the overall information is formulated as a quadratic function of variable $X$ as: 
\begin{equation}
\small
    \max_{\mathbf{X} \in \{0,1\}^N} \sum_{z\in \mathbf{D}} \mathbf{X}_z I(z) - \alpha\cdot \sum_{z, s\in \mathbf{D}} \mathbf{K}_{z, s} \mathbf{X}_z\mathbf{X}_s, ~~~~~~\text{s.t.}~~~~\sum_{z\in \mathbf{D}} \mathbf{X}_z = p, 
    \label{eq: quadratic}
\end{equation}
where $p=(1-\delta)N$ is the size budget of the selected set. In the formulation, the first-order term $I(z)$ measures the intra-sample information of the sample $z$, which can be instantiated using any existing sample-wise scores, \textit{e.g.}, EL2N \citep{GraNd_EL2N} or SSP \citep{SSP}. 
The quadratic term $\mathbf{K}_{z, s}$ measures the inter-sample redundancy between the sample $z$ and $s$.  

\paragraph{Objective Construction} 

For the intra-sample information $\mathbf{I} \in \mathbb{R}^N$ and the pairwise feature similarity  $\mathbf{\mathbf{K}} \in \mathbb{R}^{N \times N}$, 
we introduce two calculation modes by following \cite{d2pruning}, namely the supervised mode and the unsupervised mode. 

 \emph{The supervised mode.} We initially train a surrogate model on the dataset. Subsequently, we employ it to calculate sample-wise scores, such as the loss value \citep{entropy} and gradient norm \citep{GraNd_EL2N}. Additionally, we use features before the classification layer as features $\mathbf{F}^{N\times d}$, where $d$ is the feature dimension.
We utilize the inner product as the pairwise feature similarity, that is, $\mathbf{\mathbf{K}} = \mathbf{F}^\mathrm{T} \mathbf{F}$. However, the supervised mode is somewhat unfriendly as training an additional model incurs a non-negligible expense, especially on large-scale datasets. 

\emph{The unsupervised mode.} We extract features using existing open-source models such as DINO \citep{oquab2023dinov2}. Additionally, we use the inner product as the pairwise feature similarity. For the intra-sample information, we employ the SSP score \citep{SSP}. Specifically, it entails first conducting clustering in the feature space. The distance between a sample and the corresponding cluster center constitutes the SSP score.

\paragraph{Analysis} If we only preserve the first-order term by setting $\alpha$ as a very small value that is near zero, InfoMax will degenerate into a vanilla score-based scheme, that is, those samples with the highest scores will be selected. 
If we discard the first-order term by setting $\alpha$ as a very huge value, it would degenerate into the problem of finding a coreset $\mathbf{S}\in\mathbf{D}$ to minimize $\sum_{z, s\in \mathbf{S}} \mathbf{K}_{z, s}$. 
In other words, it is to find a subset with the minimum similarity within the set. 
This is a variant case of the classic k-Median problem \citep{lloyd1982least,tan2006cluster,harpeled2004on}, that minimizing the $\text{Cost}(\mathbf{S}, \mathbf{D}) = \sum_{s\in \mathbf{S}} \sum_{z\in \mathbf{D/S}} d(z, s)$ if we set the distance measurement as $d(z, s) = 1 - \mathbf{K}_{z, s}$.

\subsection{Salable InfoMax Solver} 
\label{sec:Implementation Details}

Solving the quadratic problem defined in Eq~.\ref{eq: quadratic} directly can be extremely computationally burdensome and may even prove intractable since the budget size $p$ is generally on the order of tens of thousands or even millions \citep{crf2,CRF1}. 
Here, we propose a continuous relaxation of the problem, enabling the use of a gradient-based solver for more efficient optimization. 
Firstly, we conduct convex relaxation on the feasible domain of the original optimization problem by relaxing the binary constraint $\mathbf{X} \in \{0,1\}^N$  to a continuous version $\mathbf{X} \in [0,1]^N$. 
Then, we derive a solver based on the proximal gradient method \citep{crf3,tan2021proxy}. This solver decomposes the original complex and non-convex continuous problem into a series of simple and convex sub-problems, see Appendix \ref{sec: proof} for details. Each sub-problem has an analytic solution that serves as the update rule of our InfoMax solver as follows: 
\begin{equation}
\small
\mathbf{X}^{t+1} = \text{Softmax}\Big(p\cdot\mathbf{I} - 2p\alpha\cdot \mathbf{\mathbf{K}} \mathbf{X}^{t} \Big), \label{eq:update rule}
\end{equation} 
where $\mathbf{I} \in \mathbb{R}^{N}$ is the vectorized $I(z)$ that $\mathbf{I}_z = I(z)$, and $\mathbf{\mathbf{K}} \in \mathbb{R}^{N\times N}$ is the similarity matrix. 
$\text{Softmax}$ is the operation to map the real-valued vector $\widetilde{\mathbf{X}}$ to a non-negative vector with the summation of 1, specifically, $\text{Softmax}(\mathbf{X}) = \exp(\mathbf{X})/\sum_i\exp(\mathbf{X})_i$. After $T$ iterations, we get the final solution $\mathbf{X}^{T}$, of which each item represents the probability that the corresponding sample should be selected. 
Finally, we take the coreset corresponding to the top-$p$ largest values. 

We have summarized InfoMax in Algorithm \ref{alg}. First, the algorithm requires the input of the intra-sample information vector $\mathbf{I} \in \mathbb{R}^N$ and the pairwise feature similarity  $\mathbf{\mathbf{K}} \in \mathbb{R}^{N \times N}$. Then, it will iteratively execute the iterative solver defined in Eq.~(\ref{eq:update rule}) for $T$ iterations. Next, we present additional design enhancements to further improve efficiency.  

\paragraph{Efficiency Enhancement Technique} 

Naively applying the InfoMax solver in Eq.~(\ref{eq:update rule}) on the original dataset leads to a quadratic complexity of $\mathcal{O}(N^2)$, where $N$ is the number of samples to be processed. This complexity is rather high when dealing with large-scale datasets. Here, we introduce two practical techniques to boost the efficiency of InfoMax.

\emph{Dataset partition:} Before executing the InfoMax algorithm, we divide the original dataset into $d$ smaller random subsets and then conduct pruning on each subgroup independently. With this scheme, the complexity of the algorithm on each subset is significantly reduced to $\mathcal{O}(N^2/d^2)$. At the same time, pruning for each subset can be carried out simultaneously on multiple computing devices, further reducing the time consumption. 

\emph{Sparsification:} 
Since the information overlap between distant samples is minimal, we further improve computational efficiency by sparsifying the similarity matrix $\mathbf{K}$, retaining only the top $k$ values as non-zero and setting the rest to zero. 
Specifically, we only take into account the similarity between each sample and its $k$ nearest neighbors (for instance, $k = 5$). As a result, $\mathbf{K}$ has only $Nk$ non-zero elements. With the sparsification technique, the complexity of the algorithm on each subset is significantly reduced to $\mathcal{O}(Nk/d^2)$. 

Note that the partition factor $d$ and the sparsification rate $k$ are two hyperparameters that determine the trade-off between efficiency and performance. In experiments (Sec. \ref{sec: ablation}), we also study the effects of these two hyperparameters.

\subsection{How does InfoMax find the most informative coreset?} 
\label{sec: How does InfoMax optimize the most informative coreset?}

We explain why InfoMax can find the most informative coreset from the perspective of information theory. First, we restate the information maximization formulation of data pruning defined in Eq.~(\ref{eq:true info}), 
$\mathbf{S}^{*} = \arg\max_{\mathbf{S} \subset D, |\mathbf{S}|=p} I(\mathbf{S})$, 
where $p$ is the size of the target coreset. The $I(\mathbf{S})$ measures the set-level information of the candidate subset $\mathbf{S}$, specifically, 
\begin{equation}
\small
I(\mathbf{S}) = I(z_1) + I(z_2|z_1) +  ... + I(z_p |z_1, ..., z_{p-1})= I(z_1) + \sum_{2 \leq k}^{p} I(z_k |z_1, ..., z_{k-1}). 
\label{eq: information 2}
\end{equation}
where $\{z_1, ..., z_p\} \in \mathbf{S}$ are all samples from the set. Note that this equation always holds regardless of the order of samples. 
The intra-sample information $I(z_1)$ of the sample $z_1$ could be instantiated by various sample-wise scores \citep{SSP,entropy,tan2023data,GraNd_EL2N}.  
Regarding the conditional gain term $I(z|Z)$, we refer to the recent progress in submodular information measures (SIM) \cite{Submodularity_1,Submodularity_2,Submodularity_3}, which present several instantiations for the submodular conditional gain. Here, we select the simplest yet effective instantiation, Graph Cut conditional gain (GCCG), $I(z_k |z_1, ..., z_{k-1})=f(z_k) - 2\lambda \sum_{i<k} \mathbf{K}_{z_i,z_k}$. 
It measures the dissimilarity between the sample $z_k$ and all conditional samples. Specifically, $\mathbf{K}_{z_i,z_k}$ measures the similarity between the sample $i$ and the sample $k$, $\lambda$ is an undetermined coefficient and is a hyperparameter in the system. The submodular function $f$ maps from a ground set to a real value, and we can simply set it with  $f(z)={I}(z)$. Hence, we have the following instantiation for the conditional gain term: $I(z_k |z_1, ..., z_{k-1})= I(z_k) - 2 \lambda \sum_{i<k} \mathbf{K}_{z_i,z_k}$. 
By substituting it into the Eq.~(\ref{eq: information 2}), we can instantiate and reformulate
the set-level information as: 
$I(\mathbf{S}) = \sum_{z\in\mathbf{D}} I(z) - 2\lambda\sum_{z\neq s\in\mathbf{D}} \mathbf{K}_{z,s}$. 
We introduce a set of binary variables $\mathbf{X} \in \{0, 1\}^N$ where $N=|\mathbf{D}|$ is the size of the whole training set. 
In the selection procedure, $\mathbf{X}_z=1$ indicates the sample $z$ was selected, otherwise it was pruned. 
By the problem definition in Eq.~(\ref{eq:true info}) and the set-level information formulation Eq.~(\ref{eq: information 2}), we can transform the original information maximization problem in Eq.~(\ref{eq:true info}) into the following combinatorial optimization problem, 
\begin{equation}
\small
    \max_{\mathbf{X}\in\{0,1\}^N,~|\mathbf{X}|=p}:   \sum_{z\in\mathbf{D}}  I(z) \cdot \mathbf{X}_z  - \frac{\lambda}{p-1}\sum_{z\neq s \in \mathbf{D}} \mathbf{K}_{z, s} \cdot \mathbf{X}_z \mathbf{X}_s ,
\label{eq:target}
\end{equation}
where $\lambda$ is a flexible hyperparameter to be determined in Eq.~(\ref{eq: information 2}). 
Hence, if we set $\alpha = \frac{\lambda}{p-1}$, 
then we obtain the quadratic programming problem as defined in InfoMax. Therefore, we have proved that under the premise of using the instantiation of Graph-cut conditional gain (GCCG) for the conditional gain term, solving the data pruning problem in Eq.~(\ref{eq:true info}) to find the most informative subset is equivalent to solving the quadratic problem defined in Eq.~(\ref{eq: quadratic}). See Appendix \ref{sec: proof how info max} for the proof.

\section{Experiments}
\label{sec: experiments}

We carried out extensive experiments on three tasks, namely image classification, multi-modality pretraining, and instruction tuning for Large Language Models (LLMs), to investigate the performance of our InfoMax. Subsequently, we conducted ablation studies to explore the component design within InfoMax. \textit{Each result of InfoMax is the average of five independent runs. 
The standard deviation corresponding to each result of InfoMax is less than 0.85.} 

\subsection{Image Classification}

The image classification task encompasses experiments on three datasets, namely CIFAR-10, CIFAR-100 \citep{CIFAR}, and Imagenet-1K \citep{imagenet}. Following coreset selection, we will train a model on the chosen subset to examine its performance as the performance of the coreset. The model employed here is ResNet-18 for CIFAR and ResNet-34 for ImageNet. 
For InfoMax, the dataset partition scheme is not employed herein as the dataset scale is not large. The sparse-rate $k$ is set to 5, the pairwise term weight $\alpha$ is set as 0.3, and the iteration $T$ is set as 20. Regarding the detailed experimental settings, please refer to the Appendix.

For the \textit{supervised} setting, we compare InfoMax with several baselines: 1) \textbf{Random} selection of examples. 2) \textbf{Entropy} \citep{entropy} of the model's prediction. 3) \textbf{Forgetting} \citep{forgetting} score for each example i.e., the number of times a model predicts the example incorrectly after having predicted correctly in the previous epoch. 4) \textbf{EL2N} \citep{GraNd_EL2N}, the L2 norm of error vectors. 5) \textbf{Moderate} coresets \citep{Moderate} that selects samples at about the median distance from the class center, 6) \textbf{CCS} (\textbf{CCS}) divides a range of difficulty scores into equal-sized bins and randomly samples from each bin. 
7) \textbf{$\mathbf{D^2}$-Pruning} \citep{d2pruning} selects samples by performing message passing on the data graph. 
The scoring model for each method is a ResNet model trained on the target dataset. \textbf{K-center} \citep{k_center} the standard geometry-based coreset method.  

For the \textit{Unsupervised} setting, we select the following baseline: 
1) \textbf{SSP} \citep{SSP} that uses self-supervised embeddings to compute k-means clusters and treats samples at a farther distance from the cluster center as more important, 
2) \textbf{CCS over SSP scores}, 
and 3) \textbf{$D^2$-Pruning over SSP scores} coreset selection. 
The unsupervised feature used here is from the officially public DINO-ViT-Base-V2 model \citep{oquab2023dinov2}.

\begin{table}[tp]
\caption{\label{tab:cls_results} A comparative analysis of the performance ((Acc@1)) of InfoMax on image classification datasets with ResNet-18 (on CIFAR) and ResNet-34 (on ImageNet). The best results are bolded.}
\vspace{-0.2cm}
\setlength{\tabcolsep}{3.1pt}
\centering
\resizebox{\linewidth}{!}{  
\begin{tabular}{l|cccccc|cccccc|cccccc}
    \toprule[1pt]
    \textbf{Dataset} & \multicolumn{6}{c|}{\textbf{CIFAR-10}} & \multicolumn{6}{c|}{\textbf{CIFAR-100}} & \multicolumn{6}{c}{\textbf{ImageNet-1K}} \\
    \cmidrule(lr){1-7} \cmidrule(lr){8-13} \cmidrule(lr){14-19}
    \textbf{Selection Rate} & \textbf{100\%} & \textbf{70\%} & \textbf{50\%} & \textbf{30\%} & \textbf{20\%} & \textbf{10\%} & \textbf{100\%} & \textbf{70\%} & \textbf{50\%} & \textbf{30\%} & \textbf{20\%} & \textbf{10\%} & \textbf{100\%} & \textbf{70\%} & \textbf{50\%} & \textbf{30\%} & \textbf{20\%} & \textbf{10\%} \\
    \midrule[1pt]
    \textbf{Random} & 95.5 & 94.3 & 93.4 & 90.9 & 88.0 & 79.0 & 78.7 & 74.6 & 71.1 & 65.3 & 57.4 & 44.8 & 73.1 & 72.2 & 70.3 & 66.7 & 62.5 & 52.3 \\
    \midrule[0.1pt]
    \textbf{Entropy}& - & 94.8 & 92.9 & 90.1 & 84.1 & 72.1 & - & 74.7 & 68.9 & 60.3 & 49.6 & 35.0 & - & 72.3 & 70.8 & 64.0 & 55.8 & 39.0 \\
    \textbf{K-center}& - & 94.1 & 92.5 & 91.0 & 82.5 & 68.4 & - & 73.4 & 69.5 & 61.4 & 47.2 & 40.5 & - & 73.0 & 71.4 & 64.8 & 53.1 & 42.0 \\
    \textbf{Forgetting}   & - & {95.7} & 94.9 & 88.1 & 73.8 & 46.3 & - & 76.0 & 68.1 & 49.3 & 30.3 & 20.6 & - & {72.6} & {70.9} & 66.5 & 62.9 & 52.3 \\
    \textbf{EL2N}  & - & 95.4 & 94.8 & 89.2 & 78.6 & 30.3 & - & 75.6 & 68.1 & 47.2 & 24.8 & 11.8 & - & 72.2 & 67.2 & 48.8 & 31.2 & 12.9 \\
    \textbf{SSP}  &-  &93.8  &93.1 &81.2 &64.4 &42.9  
    &- &76.5 &66.4 &50.3 &27.9 &19.4            
    &- &71.1 &68.5 &52.7 &40.3 &20.5            \\
    \textbf{Moderate}  & - & 93.9 & 92.6 & 90.6 & 87.3 & 81.0 & - & 74.6 & 71.1 & 65.3 & 58.5 & 45.5 & - & 72.0 & 70.3 & 65.9 & 61.3 & 52.1 \\
    \textbf{CCS} (unsupervised) & - &95.2 &93.4 &90.6 &85.5 &80.6 
    & - &76.4 &71.8 &61.2 &37.5 &25.4
    & - &71.6 &69.5 &62.1 &47.4 &30.2   \\
    \textbf{CCS}   & - & 95.4 & \textbf{95.0} & {93.0} & 91.0 & {86.9} & - & 77.1 & 74.4 & 68.9 & 64.0 & {57.3} & - & 72.3 & 70.5 & 67.8 & {64.5} & {57.3} \\
    \textbf{$\mathbf{D^2}$-Pruning} (unsupervised)  & - & 94.4 & 94.2 & 87.6 & 86.1 & 83.9 
    & - & 77.8 & 70.0 & 66.6 & 62.3 & 51.7
    & - & 72.6 & 69.4 & 66.1 & 60.9 & 52.5\\
     \textbf{$\mathbf{D^2}$-Pruning} & - & \textbf{95.7} & {94.9} & {93.3} & {91.4} & {87.1} 
     & - & {78.2} & {75.9} & {70.5} & {65.2} & 56.9 
     & - & {72.9} & {71.8} & {68.1} & {65.9} & 55.6 \\
     \midrule[0.1pt]
    \textbf{InfoMax} (unsupervised)  & - & {94.9} & 93.6 & {92.2} & {90.1} & {87.0}
    & - & {77.9} & {74.6} & {70.1} & {64.7} & {56.2}
    & - & {72.8} & {70.5} & {68.0} & {64.9} & {56.8}\\
    \textbf{InfoMax}   & - & {95.5} & 94.7 & \textbf{94.1} & \textbf{92.7} & \textbf{89.1}
    & - & \textbf{79.0} & \textbf{76.7} & \textbf{71.5} & \textbf{67.9} & \textbf{58.7}
    & - & \textbf{73.3} & \textbf{72.8} & \textbf{69.4} & \textbf{66.5} & \textbf{59.0}\\
\bottomrule[1pt]
\end{tabular}
}
\vspace{-0.45cm}
\end{table}

Table \ref{tab:cls_results} presents the results on three image classification datasets comparing the performance (accuracy) of InfoMax with several baselines. Firstly, let's focus on the performance in the supervised setting. On CIFAR-10, at various pruning rates (30\% - 90\%), InfoMax outperforms other methods in terms of accuracy in most settings. For instance, at a 90\% pruning rate, InfoMax achieves an accuracy of 89.1, outperforming other methods (such as $D^2$-Pruning) by 2.0\% in accuracy. On CIFAR-100 and ImageNet, InfoMax has higher accuracy values compared to other methods across different pruning rates.
Moreover, this advantage becomes more pronounced under a high pruning ratio. For example, at a 90\% pruning rate, InfoMax surpasses the second-ranked CCS by 1.4\% and 1.7\% on the two datasets respectively, and outperforms the third-ranked $D^2$-Pruning by 1.8\% and 3.4\% on the two datasets respectively. Next, we must highlight the performance of InfoMax under the unsupervised setting. The performance of InfoMax in the unsupervised setting steadily exceeds that of other schemes in the unsupervised setting. Even under the setting of a high pruning ratio, it can approach and even outperform most supervised schemes. For example, at a 90\% pruning rate, the unsupervised InfoMax on CIFAR-10 lags only 0.1\% in performance compared with the supervised $D^2$-Pruning, and the unsupervised InfoMax on ImageNet leads the supervised $D^2$-Pruning by a performance advantage of 1.2\%. 
It showcases that InfoMax exhibits superior performance and notable effectiveness across different settings.

\subsection{Multi-modality pre-training}

For multi-modality pretraining tasks, we conducted experiments on the popular vision-language dataset CC12M \citep{CC12M}, which contains 12 million image-text pairs from the Internet, for CLIP-like vision-language pre-training \citep{CLIP}. 
A common practice \citep{LAION,datacomp} for selecting coreset from VL datasets is to use the pre-trained CLIP model \citep{CLIP} to score each image-text pair, where a higher CLIP score indicates better image-text alignment. 
Hence, we set this as the most basic baseline, termed, 1) \textbf{CLIP score}. 
Additionally, we also select the following baseline: 
2) \textbf{Clustering + CLIP}: Since using only the score for screening is based on the matching degree of images and texts it is difficult to reflect the redundancy degree of samples. Consequently, some schemes \citep{li2022blip,li2023blip2} combine Clustering and CLIP scores, that is, first clustering in the feature space and then selecting a portion of samples with high scores within each cluster. 
3) \textbf{Moderate} coreset \citep{Moderate} over CLIP scores: Selecting those samples with the median score since the highly-scored samples may be the too-easy samples.  
4) \textbf{CCS} over CLIP scores: By following the implementation in \cite{d2pruning}, it divides a range of CLIP scores into equal-sized bins and randomly samples from each bin. 
5) \textbf{$\mathbf{D^2}$-Pruning} over CLIP scores: It constructs a sample graph by using the CLIP score as the node value and using image features from the CLIP vision encoder to calculate the edge value (similarity).

For our InfoMax approach, we also designate the CLIP score as the intra-sample information measurement. For the pairwise similarity, we employ the inner product between the features of samples from the CLIP vision encoder. The dataset partition factor $d$ is set to 10; that is, we randomly partition the 12M data into 10 subsets each of size 1.2M. The sparse-rate $k$ is set at 5, the pairwise term weight $\alpha$ is set as 0.3, and the iteration $T$ is set to 20.   
After data selection, we perform CLIP pre-training on the coreset and evaluate the trained model on four downstream tasks: Zero-shot ImageNet-1K Classification, Linear Probing ImageNet-1K Classification, Image-to-Text (I2T) Flickr30K \citep{flickr30k} Retrieval and Text-to-Image (T2I) Flickr30K Retrieval. For the detailed experimental settings and results along with standard errors, please refer to the Appendix.

\begin{figure*}[tp]
\centering 
\includegraphics[width=1\linewidth]{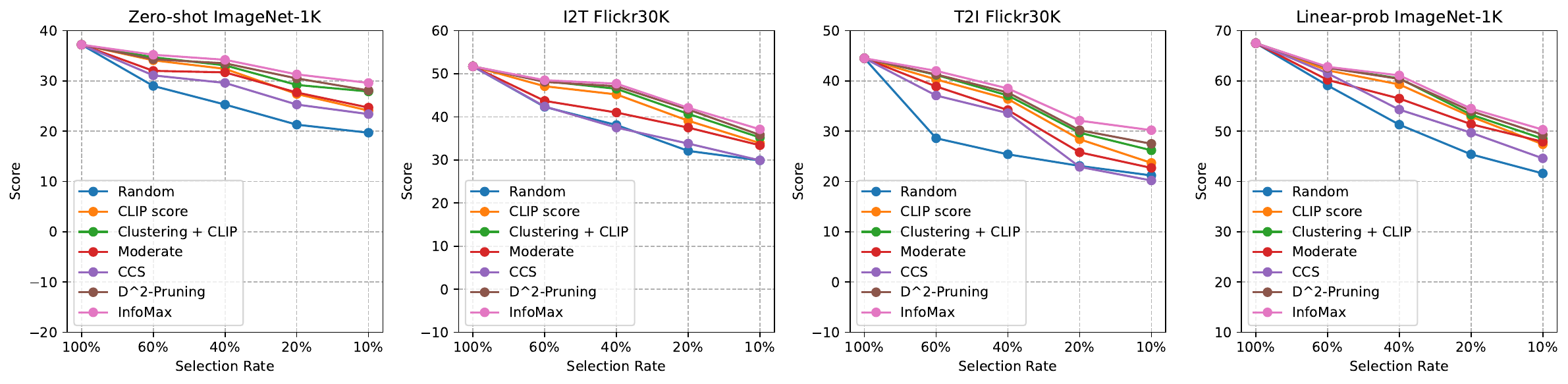}
\caption{\label{tb: coreset clip} 
Experimental results of coreset selection on CC12M \citep{CC12M} for multi-modality (vision-language) pre-training on CLIP-ViT-B/32 \citep{CLIP} model. 
} 
\vspace{-0.3cm}
\end{figure*}

The experimental results are presented in Table \ref{tb: coreset clip}. InfoMax consistently outperforms all baselines across all selection ratios. This leading advantage is particularly prominent when the selection ratio is relatively small. For instance, when selection ratios = 10\%, InfoMax surpasses $D^2$-Pruning by 1.5\%, 1.3\%, 2.7\%, and 1.0\% respectively on these four downstream tasks, and exceeds the widely-used selection method of Clustering + CLIP by 2.1\%, 1.9\%, 4.0\%, and 1.8\% respectively on the four tasks. This demonstrates the superiority and effectiveness of InfoMax in different scenarios and under various selection conditions, highlighting its significance in data-centric research areas.

\subsection{Instruction Tuning for Large Language Models}

Recently, some works \citep{xia2024less, d2pruning} have demonstrated significant redundancy in language datasets. For instance, LESS \citep{xia2024less} reduced the size of instruction tuning datasets to merely 5\% of their original amount. The core of LESS is the influence score \citep{TracIN, tan2023data}, which gauges how a particular data point impacts the performance of the learned model on the validation set. 

Following \citep{xia2024less}, we also conduct coreset selection on a mixed dataset containing data from FLAN-V2~\citep{longpre2023flan}, COT~\citep{wei2022chain},  DOLLY~\citep{DatabricksBlog2023DollyV2}, OPEN-ASSISTANT-1~\citep{kopf2023openassistant}. 
The mixed training set, consisting of approximately 270K data points, exhibits a wide variation in its format and the underlying reasoning tasks. 
After the selection process, we conduct LoRA-finetuning \citep{hu2021lora} on the selected coreset for a LLaMA2-7B model \citep{touvron2023llama2}. 
For $D^2$-Pruning \citep{d2pruning} and our InfoMax, we employ the LESS score as the intra-sample measurement and utilize the gradient of the sample as the sample's feature. Additionally, we select Moderate \citep{Moderate} over LESS and CCS \citep{CCS} over LESS as the baselines.

\begin{figure*}[thtp]
\centering 
\includegraphics[width=0.8\linewidth]{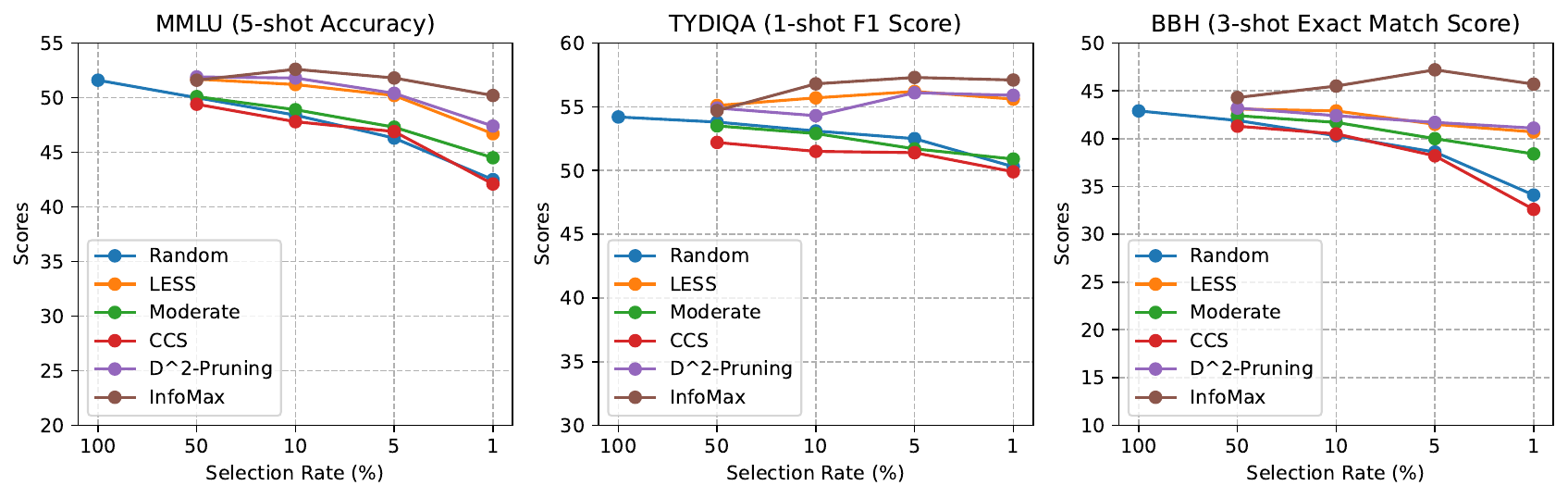}
\caption{\label{tab: nlp_results} A comparative analysis of the performance of InfoMax on instruction tuning datasets for LLaMA2-7B \citep{touvron2023llama2} model. 
} 
\end{figure*}

After the training process, we assess our method using three widely recognized benchmarks: 1) MMLU \citep{hendrycks2020measuring} offers a diverse range of knowledge domains for evaluation, covering 57 knowledge areas, like mathematics, computer science, and others. 
2) TYDIQA~\citep{tydiqa} is a multilingual question-answer dataset that includes 9 kinds of languages. 
3) BBH~\citep{suzgun2023challenging} encompasses 27 arduous tasks to assess whether the model can handle complex reasoning situations. 
The experimental results are presented in Figure~\ref{tab: nlp_results}. 
InfoMax, our proposed method, demonstrates significant performance advantages over other competing methods, particularly at small selection rates ($\leq 5\%$). On MMLU, it shows notable improvements compared to the second-ranked method, $D^2$-Pruning. For example, at the 95\% and 99\% pruning rates, InfoMax achieves an accuracy of 51.8 and 50.2, compared to $D^2$-Pruning's performance, an improvement of about 2\% in performance. Similar trends are observed on TYDIQA, InfoMax outperforms LESS (the second-ranked approach) at 90\%, 95\%, and 99\% pruning rates, with differences ranging from about 1.1 to 1.5 percentage points. For the BBH dataset, the superiority of InfoMax is even more pronounced. At a 95\% pruning rate, it has an accuracy of 47.2 compared to $D^2$-Pruning's 41.7, a substantial improvement of about 5.5 percentage points. Overall, InfoMax consistently shows better results, highlighting its effectiveness even when a large portion of the data is pruned.

\subsection{Analysis \& Discussion}
\label{sec: ablation}

\begin{figure*}[http]
\centering 
\vspace{-0.2cm}
\includegraphics[width=1\linewidth]{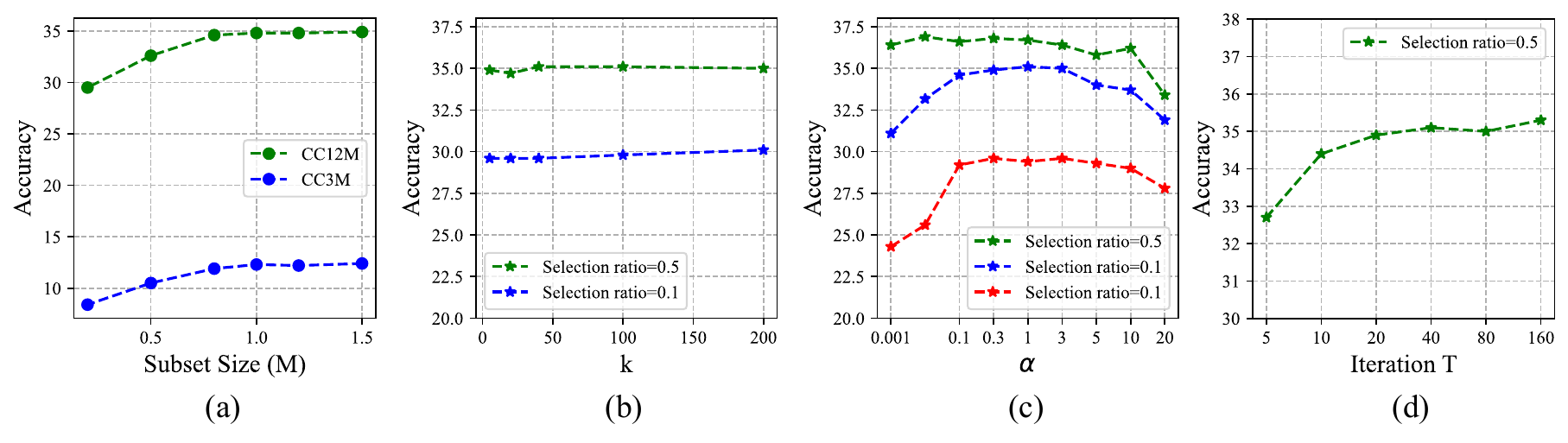}
\vspace{-0.8cm}
\caption{\label{fig: ablation}
Ablation study for hyperparameters in InfoMax: the dataset partition strategy (the size of each subset), the sparse rate $k$, the weight $\alpha$ of the pairwise term, and the iteration number $T$ of the InfoMax solver. Experiments in (a) are conducted on CC3M \& CC12M \citep{CC12M}. All experiments in (b,c,d) are conducted on CC12M. The selected model is CLIP-ViT-B/32 \citep{CLIP}. The reported metric focuses on the accuracy of the coreset-trained CLIP-ViT-B/32 on the zero-shot ImageNet-1K classification tasks.
}
\vspace{-0.2cm}
\end{figure*}

Here, we have investigated four hyperparameters in InfoMax: the dataset partition rate $d$, the sparse rate $k$, the weight $\alpha$ of the pairwise term, and the iteration $T$ for InfoMax. In Appendix Sec.~\ref{sec: Further Analysis}, we will study the time cost and generalization ability of InfoMax. 
\textbf{(a) Partition strategy.} The experiments on CC3M/CC12M \citep{CC12M} in Figure \ref{fig: ablation}(a) studies the effect of the dataset partition strategy. It suggests that a decrease in the partitioned subset size inevitably reduces the performance of the coreset. This decline exhibits a significant downward tendency when the size of each subset is less than approximately 1M. Consequently, when dealing with relatively large-scale datasets, we recommend ensuring that the size of each subset is at least 1M samples. $~~$
\textbf{(b) Sparse rate $k$:} The sparse rate $k$ represents the size of the neighborhood of each sample. For details, see Sec.~\ref{sec:Implementation Details}. A smaller $k$ implies more 0 items in the similarity matrix $K$. 
In Figure \ref{fig: ablation}(b), we found that the performance of InfoMax is relatively robust to the choice of $k$. Hence, we set $k = 5$ for better efficiency in all experiments. 
$~~$\textbf{(c) Pairwise weight $\alpha$:} 
The pairwise weight $\alpha$ determines the tradeoff between the first-order term and the second-order term in the formulation of InfoMax as in Eq.\ref{eq: quadratic}.
As shown in Figure \ref{fig: ablation}(c), it is observed that the optimal range of $\alpha$ varies for different selection ratios. However, all these ranges yield satisfactory results when $\alpha$ is around 0.3 to 4. Consequently, we recommend setting $\alpha = 0.3$. 
$~~$\textbf{(d) Iteration $T$:} Figure \ref{fig: ablation}(d) reveals that the performance of InfoMax rises with the increase in the iteration number $T$. Nevertheless, this gain tends to saturate after $T > 20$. Hence, we suggest setting $T = 20$ to balance good performance and efficiency. 

\section{Conclusion}

This paper has presented InfoMax, a novel and effective data pruning method. InfoMax is formulated as a quadratic optimization problem to maximize the sample-wise informativeness and minimize the redundancy of selected samples. Furthermore, an efficient gradient-based solver along with sparsification techniques and dataset partitioning strategies is introduced to ensure scalability, enabling it to handle datasets with millions of samples within tens of minutes. Overall, InfoMax shows great potential and effectiveness in the field of data pruning to improve the efficiency and performance of data processing in various applications. 

\section*{Acknowledgment} 
This work has been supported in part by the Hong Kong Research Grant Council - Early Career Scheme (Grant No. 27209621), General Research Fund Scheme (Grant No. 17202422, 17212923), Theme-based Research (Grant No. T45-701/22-R), and the Shenzhen Science and Technology Innovation Commission (SGDX20220530111405040).  Part of the described research work is conducted in the JC STEM Lab of Robotics for Soft Materials funded by The Hong Kong Jockey Club Charities Trust.

\bibliography{main}

\begin{thebibliography}{64}
\providecommand{\natexlab}[1]{#1}
\providecommand{\url}[1]{\texttt{#1}}
\expandafter\ifx\csname urlstyle\endcsname\relax
  \providecommand{\doi}[1]{doi: #1}\else
  \providecommand{\doi}{doi: \begingroup \urlstyle{rm}\Url}\fi

\bibitem[Ash et~al.(2020)Ash, Zhang, Krishnamurthy, Langford, and
  Agarwal]{BADGE}
Jordan~T. Ash, Chicheng Zhang, Akshay Krishnamurthy, John Langford, and Alekh
  Agarwal.
\newblock Deep batch active learning by diverse, uncertain gradient lower
  bounds.
\newblock In \emph{International Conference on Learning Representations}, 2020.

\bibitem[Baqué et~al.(2016)Baqué, Bagautdinov, Fleuret, and Fua]{crf3}
Pierre Baqué, Timur Bagautdinov, François Fleuret, and Pascal Fua.
\newblock Principled parallel mean-field inference for discrete random fields.
\newblock In \emph{2016 IEEE Conference on Computer Vision and Pattern
  Recognition (CVPR)}, pp.\  5848--5857, 2016.

\bibitem[Brown(2020)]{GPT3}
Tom~B Brown.
\newblock Language models are few-shot learners.
\newblock 2020.

\bibitem[Chan et~al.(2022)Chan, Yu, You, Qi, Wright, and Ma]{chan2022redunet}
Kwan Ho~Ryan Chan, Yaodong Yu, Chong You, Haozhi Qi, John Wright, and Yi~Ma.
\newblock Redunet: A white-box deep network from the principle of maximizing
  rate reduction.
\newblock \emph{The Journal of Machine Learning Research}, 23\penalty0
  (1):\penalty0 4907--5009, 2022.

\bibitem[Changpinyo et~al.(2021)Changpinyo, Sharma, Ding, and Soricut]{CC12M}
Soravit Changpinyo, Piyush Sharma, Nan Ding, and Radu Soricut.
\newblock Conceptual 12m: Pushing web-scale image-text pre-training to
  recognize long-tail visual concepts.
\newblock In \emph{Proceedings of the IEEE/CVF Conference on Computer Vision
  and Pattern Recognition}, pp.\  3558--3568, 2021.

\bibitem[Clark et~al.(2020)Clark, Choi, Collins, Garrette, Kwiatkowski,
  Nikolaev, and Palomaki]{tydiqa}
Jonathan~H. Clark, Eunsol Choi, Michael Collins, Dan Garrette, Tom Kwiatkowski,
  Vitaly Nikolaev, and Jennimaria Palomaki.
\newblock {TyDi QA}: A benchmark for information-seeking question answering in
  typologically diverse languages.
\newblock \emph{Transactions of the Association for Computational Linguistics},
  2020.

\bibitem[Coates \& Ng(2012)Coates and Ng]{coates2012learning}
Adam Coates and Andrew~Y. Ng.
\newblock Learning feature representations with $k$-means.
\newblock In \emph{Neural Networks: Tricks of the Trade - Second Edition}, pp.\
   561--580. 2012.

\bibitem[{Cody Coleman} et~al.(2019){Cody Coleman}, {Christopher Yeh}, {Stephen
  Mussmann}, {Baharan Mirzasoleiman}, {Peter Bailis}, {Percy Liang}, {Jure
  Leskovec}, and {Matei Zaharia}]{entropy}
{Cody Coleman}, {Christopher Yeh}, {Stephen Mussmann}, {Baharan Mirzasoleiman},
  {Peter Bailis}, {Percy Liang}, {Jure Leskovec}, and {Matei Zaharia}.
\newblock Selection via proxy: Efficient data selection for deep learning.
\newblock In \emph{International Conference on Learning Representations}, 2019.

\bibitem[Conover et~al.(2023)Conover, Hayes, Mathur, Xie, Wan, Shah, Ghodsi,
  Wendell, Zaharia, and Xin]{DatabricksBlog2023DollyV2}
Mike Conover, Matt Hayes, Ankit Mathur, Jianwei Xie, Jun Wan, Sam Shah, Ali
  Ghodsi, Patrick Wendell, Matei Zaharia, and Reynold Xin.
\newblock Free {Dolly}: Introducing the world's first truly open
  instruction-tuned {LLM}, 2023.

\bibitem[Esser et~al.(2020)Esser, Rombach, and Ommer]{vqgan}
Patrick Esser, Robin Rombach, and Björn Ommer.
\newblock Taming transformers for high-resolution image synthesis, 2020.

\bibitem[Feldman \& Langberg(2011)Feldman and Langberg]{feldman2011unified}
Dan Feldman and Michael Langberg.
\newblock A unified framework for approximating and clustering data.
\newblock In \emph{Proceedings of the forty-third annual ACM symposium on
  Theory of computing}, pp.\  569--578. ACM, 2011.

\bibitem[Feldman et~al.(2013)Feldman, Schmidt, and Sohler]{feldman2013turning}
Dan Feldman, Melanie Schmidt, and Christian Sohler.
\newblock Turning big data into tiny data: Constant-size coresets for
  $k$-means, {PCA} and projective clustering.
\newblock In \emph{Proceedings of the Twenty-Fourth Annual ACM-SIAM Symposium
  on Discrete Algorithms}, pp.\  1434--1453. SIAM, 2013.

\bibitem[Gadre et~al.(2024)Gadre, Ilharco, Fang, Hayase, Smyrnis, Nguyen,
  Marten, Wortsman, Ghosh, Zhang, et~al.]{datacomp}
Samir~Yitzhak Gadre, Gabriel Ilharco, Alex Fang, Jonathan Hayase, Georgios
  Smyrnis, Thao Nguyen, Ryan Marten, Mitchell Wortsman, Dhruba Ghosh, Jieyu
  Zhang, et~al.
\newblock Datacomp: In search of the next generation of multimodal datasets.
\newblock \emph{Advances in Neural Information Processing Systems}, 36, 2024.

\bibitem[Har-Peled \& Mazumdar(2004)Har-Peled and Mazumdar]{coreset}
Sariel Har-Peled and Soham Mazumdar.
\newblock On coresets for k-means and k-median clustering.
\newblock In \emph{Proceedings of the Thirty-Sixth Annual ACM Symposium on
  Theory of Computing}, pp.\  291–300, 2004.

\bibitem[{Har-Peled} \& Mazumdar(2004){Har-Peled} and Mazumdar]{harpeled2004on}
Sariel {Har-Peled} and Soham Mazumdar.
\newblock On coresets for $k$-means and $k$-median clustering.
\newblock In \emph{36th Annual ACM Symposium on Theory of Computing,}, pp.\
  291--300, 2004.

\bibitem[Har-Peled et~al.(2007)Har-Peled, Roth, and Zimak]{margin}
Sariel Har-Peled, Dan Roth, and Dav Zimak.
\newblock Maximum margin coresets for active and noise tolerant learning.
\newblock In \emph{Proceedings of the 20th International Joint Conference on
  Artifical Intelligence}, pp.\  836–841, 2007.

\bibitem[He et~al.(2024)He, Yang, Huang, and Zhao]{Dyn_Unc}
Muyang He, Shuo Yang, Tiejun Huang, and Bo~Zhao.
\newblock Large-scale dataset pruning with dynamic uncertainty.
\newblock In \emph{Computer Vision and Pattern Recognition Workshops}, 2024.

\bibitem[Hendrycks et~al.(2020)Hendrycks, Burns, Basart, Zou, Mazeika, Song,
  and Steinhardt]{hendrycks2020measuring}
Dan Hendrycks, Collin Burns, Steven Basart, Andy Zou, Mantas Mazeika, Dawn
  Song, and Jacob Steinhardt.
\newblock Measuring massive multitask language understanding.
\newblock In \emph{International Conference on Learning Representations}, 2020.

\bibitem[Hu et~al.(2021)Hu, Shen, Wallis, Allen-Zhu, Li, Wang, Wang, and
  Chen]{hu2021lora}
Edward~J Hu, Yelong Shen, Phillip Wallis, Zeyuan Allen-Zhu, Yuanzhi Li, Shean
  Wang, Lu~Wang, and Weizhu Chen.
\newblock Lora: Low-rank adaptation of large language models.
\newblock \emph{arXiv preprint arXiv:2106.09685}, 2021.

\bibitem[Jiang et~al.(2024)Jiang, Cheng, Chen, Wang, and Wei]{Dos}
Wenyu Jiang, Hao Cheng, MingCai Chen, Chongjun Wang, and Hongxin Wei.
\newblock {DOS}: Diverse outlier sampling for out-of-distribution detection.
\newblock In \emph{The Twelfth International Conference on Learning
  Representations}, 2024.

\bibitem[{Jordan T Ash} et~al.(2019){Jordan T Ash}, {Chicheng Zhang}, {Akshay
  Krishnamurthy}, {John Langford}, and {Alekh Agarwal}]{AL5}
{Jordan T Ash}, {Chicheng Zhang}, {Akshay Krishnamurthy}, {John Langford}, and
  {Alekh Agarwal}.
\newblock Deep batch active learning by diverse, uncertain gradient lower
  bounds.
\newblock In \emph{International Conference on Learning Representations}, 2019.

\bibitem[Kaushal et~al.(2021)Kaushal, Kothawade, Ramakrishnan, Bilmes, and
  Iyer]{Submodularity_2}
Vishal Kaushal, Suraj Kothawade, Ganesh Ramakrishnan, Jeff Bilmes, and Rishabh
  Iyer.
\newblock Prism: A unified framework of parameterized submodular information
  measures for targeted data subset selection and summarization.
\newblock 2021.

\bibitem[Killamsetty et~al.(2021)Killamsetty, Durga, Ramakrishnan, De, and
  Iyer]{gradient_2}
Krishnateja Killamsetty, Sivasubramanian Durga, Ganesh Ramakrishnan, Abir De,
  and Rishabh Iyer.
\newblock Grad-match: Gradient matching based data subset selection for
  efficient deep model training.
\newblock In \emph{International Conference on Machine Learning}, pp.\
  5464--5474. PMLR, 2021.

\bibitem[Kirillov et~al.(2023)Kirillov, Mintun, Ravi, Mao, Rolland, Gustafson,
  Xiao, Whitehead, Berg, Lo, et~al.]{SAM}
Alexander Kirillov, Eric Mintun, Nikhila Ravi, Hanzi Mao, Chloe Rolland, Laura
  Gustafson, Tete Xiao, Spencer Whitehead, Alexander~C Berg, Wan-Yen Lo, et~al.
\newblock Segment anything.
\newblock pp.\  4015--4026, 2023.

\bibitem[K{\"o}pf et~al.(2023)K{\"o}pf, Kilcher, von R{\"u}tte, Anagnostidis,
  Tam, Stevens, Barhoum, Duc, Stanley, Nagyfi, et~al.]{kopf2023openassistant}
Andreas K{\"o}pf, Yannic Kilcher, Dimitri von R{\"u}tte, Sotiris Anagnostidis,
  Zhi-Rui Tam, Keith Stevens, Abdullah Barhoum, Nguyen~Minh Duc, Oliver
  Stanley, Rich{\'a}rd Nagyfi, et~al.
\newblock {OpenAssistant} conversations--democratizing large language model
  alignment.
\newblock 2023.

\bibitem[Kothawade et~al.(2021)Kothawade, Beck, Killamsetty, and
  Iyer]{Submodularity_3}
Suraj Kothawade, Nathan Beck, Krishnateja Killamsetty, and Rishabh Iyer.
\newblock Similar: Submodular information measures based active learning in
  realistic scenarios.
\newblock volume~34, pp.\  18685--18697, 2021.

\bibitem[Kr{\"a}henb{\"u}hl \& Koltun(2011)Kr{\"a}henb{\"u}hl and Koltun]{crf2}
Philipp Kr{\"a}henb{\"u}hl and Vladlen Koltun.
\newblock Efficient inference in fully connected crfs with gaussian edge
  potentials.
\newblock \emph{Advances in neural information processing systems}, 24, 2011.

\bibitem[{Kristof Meding} et~al.(2022){Kristof Meding}, {Luca M. Schulze
  Buschoff}, {Robert Geirhos}, and {Felix A. Wichmann}]{ddd}
{Kristof Meding}, {Luca M. Schulze Buschoff}, {Robert Geirhos}, and {Felix A.
  Wichmann}.
\newblock Trivial or impossible—dichotomous data difficulty masks model
  differences (on imagenet and beyond).
\newblock In \emph{International Conference on Learning Representations}, 2022.

\bibitem[Krizhevsky(2009)]{CIFAR}
Alex Krizhevsky.
\newblock Learning multiple layers of features from tiny images.
\newblock \emph{Technical report}, 2009.

\bibitem[Larsson et~al.(2018)Larsson, Arnab, Kahl, Zheng, and Torr]{CRF1}
M{\aa}ns Larsson, Anurag Arnab, Fredrik Kahl, Shuai Zheng, and Philip Torr.
\newblock A projected gradient descent method for crf inference allowing
  end-to-end training of arbitrary pairwise potentials.
\newblock In Marcello Pelillo and Edwin Hancock (eds.), \emph{Energy
  Minimization Methods in Computer Vision and Pattern Recognition}, 2018.

\bibitem[Li et~al.(2022)Li, Li, Xiong, and Hoi]{li2022blip}
Junnan Li, Dongxu Li, Caiming Xiong, and Steven Hoi.
\newblock Blip: Bootstrapping language-image pre-training for unified
  vision-language understanding and generation.
\newblock In \emph{ICML}, 2022.

\bibitem[Li et~al.(2023)Li, Li, Savarese, and Hoi]{li2023blip2}
Junnan Li, Dongxu Li, Silvio Savarese, and Steven Hoi.
\newblock {BLIP-2:} bootstrapping language-image pre-training with frozen image
  encoders and large language models.
\newblock In \emph{ICML}, 2023.

\bibitem[Lloyd(1982)]{lloyd1982least}
Stuart~P. Lloyd.
\newblock Least squares quantization in {PCM}.
\newblock \emph{{IEEE} Trans. Information Theory}, 28\penalty0 (2):\penalty0
  129--136, 1982.

\bibitem[Longpre et~al.(2023)Longpre, Hou, Vu, Webson, Chung, Tay, Zhou, Le,
  Zoph, Wei, et~al.]{longpre2023flan}
Shayne Longpre, Le~Hou, Tu~Vu, Albert Webson, Hyung~Won Chung, Yi~Tay, Denny
  Zhou, Quoc~V Le, Barret Zoph, Jason Wei, et~al.
\newblock The flan collection: Designing data and methods for effective
  instruction tuning.
\newblock \emph{arXiv preprint arXiv:2301.13688}, 2023.

\bibitem[Maharana et~al.(2023)Maharana, Yadav, and Bansal]{d2pruning}
Adyasha Maharana, Prateek Yadav, and Mohit Bansal.
\newblock D2 pruning: Message passing for balancing diversity \& difficulty in
  data pruning.
\newblock 2023.

\bibitem[Mahmoud et~al.(2024)Mahmoud, Elhoushi, Abbas, Yang, Ardalani, Leather,
  and Morcos]{Sieve}
Anas Mahmoud, Mostafa Elhoushi, Amro Abbas, Yu~Yang, Newsha Ardalani, Hugh
  Leather, and Ari~S. Morcos.
\newblock {Sieve: Multimodal Dataset Pruning Using Image Captioning Models }.
\newblock In \emph{Computer Vision and Pattern Recognition (CVPR)}, 2024.

\bibitem[Mirzasoleiman et~al.(2020)Mirzasoleiman, Bilmes, and
  Leskovec]{gradient_1}
Baharan Mirzasoleiman, Jeff Bilmes, and Jure Leskovec.
\newblock Coresets for data-efficient training of machine learning models.
\newblock In \emph{International Conference on Machine Learning}, pp.\
  6950--6960. PMLR, 2020.

\bibitem[Oquab et~al.(2023)Oquab, Darcet, Moutakanni, Vo, Szafraniec, Khalidov,
  Fernandez, Haziza, Massa, El-Nouby, Howes, Huang, Xu, Sharma, Li, Galuba,
  Rabbat, Assran, Ballas, Synnaeve, Misra, Jegou, Mairal, Labatut, Joulin, and
  Bojanowski]{oquab2023dinov2}
Maxime Oquab, Timothée Darcet, Theo Moutakanni, Huy~V. Vo, Marc Szafraniec,
  Vasil Khalidov, Pierre Fernandez, Daniel Haziza, Francisco Massa, Alaaeldin
  El-Nouby, Russell Howes, Po-Yao Huang, Hu~Xu, Vasu Sharma, Shang-Wen Li,
  Wojciech Galuba, Mike Rabbat, Mido Assran, Nicolas Ballas, Gabriel Synnaeve,
  Ishan Misra, Herve Jegou, Julien Mairal, Patrick Labatut, Armand Joulin, and
  Piotr Bojanowski.
\newblock Dinov2: Learning robust visual features without supervision, 2023.

\bibitem[Paszke et~al.(2017)Paszke, Gross, Chintala, Chanan, Yang, DeVito, Lin,
  Desmaison, Antiga, and Lerer]{pytorch}
Adam Paszke, Sam Gross, Soumith Chintala, Gregory Chanan, Edward Yang, Zachary
  DeVito, Zeming Lin, Alban Desmaison, Luca Antiga, and Adam Lerer.
\newblock Automatic differentiation in pytorch.
\newblock 2017.

\bibitem[Paul et~al.(2018)Paul, {Surya Ganguli}, and {Gintare Karolina
  Dziugaite}]{GraNd_EL2N}
Mansheej Paul, {Surya Ganguli}, and {Gintare Karolina Dziugaite}.
\newblock Deep learning on a data diet: Finding important examples early in
  training.
\newblock In \emph{Advances in Neural Information Processing Systems}, 2018.

\bibitem[Plummer et~al.(2015)Plummer, Wang, Cervantes, Caicedo, Hockenmaier,
  and Lazebnik]{flickr30k}
Bryan~A Plummer, Liwei Wang, Chris~M Cervantes, Juan~C Caicedo, Julia
  Hockenmaier, and Svetlana Lazebnik.
\newblock Flickr30k entities: Collecting region-to-phrase correspondences for
  richer image-to-sentence models.
\newblock In \emph{Proceedings of the IEEE international conference on computer
  vision}, pp.\  2641--2649, 2015.

\bibitem[Pruthi et~al.(2020)Pruthi, Liu, Mukund, and Kale]{TracIN}
Garima Pruthi, Frederick Liu, Sundararajan Mukund, and Satyen Kale.
\newblock Estimating training data influence by tracing gradient descent.
\newblock \emph{arXiv preprint arXiv:2002.08484}, 2020.

\bibitem[Radford et~al.(2021)Radford, Kim, Hallacy, Ramesh, Goh, Agarwal,
  Sastry, Askell, Mishkin, Clark, et~al.]{CLIP}
Alec Radford, Jong~Wook Kim, Chris Hallacy, Aditya Ramesh, Gabriel Goh,
  Sandhini Agarwal, Girish Sastry, Amanda Askell, Pamela Mishkin, Jack Clark,
  et~al.
\newblock Learning transferable visual models from natural language
  supervision.
\newblock pp.\  8748--8763, 2021.

\bibitem[Rombach et~al.(2022)Rombach, Blattmann, Lorenz, Esser, and Ommer]{SD}
Robin Rombach, Andreas Blattmann, Dominik Lorenz, Patrick Esser, and Bj{\"o}rn
  Ommer.
\newblock High-resolution image synthesis with latent diffusion models.
\newblock pp.\  10684--10695, 2022.

\bibitem[Russakovsky et~al.(2015)Russakovsky, Deng, Su, Krause, Satheesh, Ma,
  Huang, Karpathy, Khosla, Bernstein, et~al.]{imagenet}
Olga Russakovsky, Jia Deng, Hao Su, Jonathan Krause, Sanjeev Satheesh, Sean Ma,
  Zhiheng Huang, Andrej Karpathy, Aditya Khosla, Michael Bernstein, et~al.
\newblock Imagenet large scale visual recognition challenge.
\newblock \emph{International journal of computer vision}, 115:\penalty0
  211--252, 2015.

\bibitem[Schuhmann et~al.(2022)Schuhmann, Beaumont, Vencu, Gordon, Wightman,
  Cherti, Coombes, Katta, Mullis, Wortsman, et~al.]{LAION}
Christoph Schuhmann, Romain Beaumont, Richard Vencu, Cade Gordon, Ross
  Wightman, Mehdi Cherti, Theo Coombes, Aarush Katta, Clayton Mullis, Mitchell
  Wortsman, et~al.
\newblock Laion-5b: An open large-scale dataset for training next generation
  image-text models.
\newblock volume~35, pp.\  25278--25294, 2022.

\bibitem[Sener \& Savarese(2017)Sener and Savarese]{k_center}
Ozan Sener and Silvio Savarese.
\newblock Active learning for convolutional neural networks: A core-set
  approach.
\newblock 2017.

\bibitem[Sorscher et~al.(2022)Sorscher, Geirhos, Shekhar, Ganguli, and
  Morcos]{SSP}
Ben Sorscher, Robert Geirhos, Shashank Shekhar, Surya Ganguli, and Ari Morcos.
\newblock Beyond neural scaling laws: beating power law scaling via data
  pruning.
\newblock In \emph{Advances in Neural Information Processing Systems}, 2022.

\bibitem[Suzgun et~al.(2023)Suzgun, Scales, Sch{\"a}rli, Gehrmann, Tay, Chung,
  Chowdhery, Le, Chi, Zhou, et~al.]{suzgun2023challenging}
Mirac Suzgun, Nathan Scales, Nathanael Sch{\"a}rli, Sebastian Gehrmann, Yi~Tay,
  Hyung~Won Chung, Aakanksha Chowdhery, Quoc Le, Ed~Chi, Denny Zhou, et~al.
\newblock Challenging big-bench tasks and whether chain-of-thought can solve
  them.
\newblock In \emph{Findings of the Association for Computational Linguistics:
  ACL 2023}, pp.\  13003--13051, 2023.

\bibitem[Tan et~al.(2021)Tan, Wang, Wu, Wang, Zhang, and Liu]{tan2021proxy}
H.-R. Tan, C.~Wang, S.-T. Wu, T.-Q. Wang, X.-Y. Zhang, and C.-L. Liu.
\newblock Proxy graph matching with proximal matching networks.
\newblock \emph{Proceedings of the AAAI Conference on Artificial Intelligence},
  35\penalty0 (11):\penalty0 9808--9815, 2021.

\bibitem[Tan et~al.(2023)Tan, Wu, Du, Chen, Wang, Wang, and Qi]{tan2023data}
Haoru Tan, Sitong Wu, Fei Du, Yukang Chen, Zhibin Wang, Fan Wang, and Xiaojuan
  Qi.
\newblock Data pruning via moving-one-sample-out.
\newblock In \emph{Advances in neural information processing systems}, 2023.

\bibitem[Tan et~al.(2006)Tan, Steinbach, Kumar, et~al.]{tan2006cluster}
Pang-Ning Tan, Michael Steinbach, Vipin Kumar, et~al.
\newblock Cluster analysis: basic concepts and algorithms.
\newblock \emph{Introduction to data mining}, 8:\penalty0 487--568, 2006.

\bibitem[Toneva et~al.(2018)Toneva, {Alessandro Sordoni}, {Remi Tachet des
  Combes}, {Adam Trischler}, {Yoshua Bengio}, and {Geoffrey J
  Gordon}]{forgetting}
Mariya Toneva, {Alessandro Sordoni}, {Remi Tachet des Combes}, {Adam
  Trischler}, {Yoshua Bengio}, and {Geoffrey J Gordon}.
\newblock An empirical study of example forgetting during deep neural network
  learning.
\newblock In \emph{International Conference on Learning Representations}, 2018.

\bibitem[Touvron et~al.(2023)Touvron, Martin, Stone, Albert, Almahairi, Babaei,
  Bashlykov, Batra, Bhargava, Bhosale, et~al.]{touvron2023llama2}
Hugo Touvron, Louis Martin, Kevin Stone, Peter Albert, Amjad Almahairi, Yasmine
  Babaei, Nikolay Bashlykov, Soumya Batra, Prajjwal Bhargava, Shruti Bhosale,
  et~al.
\newblock Llama 2: Open foundation and fine-tuned chat models.
\newblock \emph{arXiv preprint arXiv:2307.09288}, 2023.

\bibitem[{Vitaly Feldman} \& {Chiyuan Zhang}(2020){Vitaly Feldman} and {Chiyuan
  Zhang}]{memory}
{Vitaly Feldman} and {Chiyuan Zhang}.
\newblock What neural networks memorize and why: Discovering the long tail via
  influence estimation.
\newblock In \emph{Advances in Neural Information Processing Systems}, 2020.

\bibitem[Wei et~al.(2022)Wei, Wang, Schuurmans, Bosma, Xia, Chi, Le, Zhou,
  et~al.]{wei2022chain}
Jason Wei, Xuezhi Wang, Dale Schuurmans, Maarten Bosma, Fei Xia, Ed~Chi, Quoc~V
  Le, Denny Zhou, et~al.
\newblock Chain-of-thought prompting elicits reasoning in large language
  models.
\newblock \emph{Advances in Neural Information Processing Systems},
  35:\penalty0 24824--24837, 2022.

\bibitem[Wei et~al.(2015)Wei, Iyer, and Bilmes]{Submodularity_1}
Kai Wei, Rishabh Iyer, and Jeff Bilmes.
\newblock Submodularity in data subset selection and active learning.
\newblock In \emph{International conference on machine learning}, pp.\
  1954--1963. PMLR, 2015.

\bibitem[Xia et~al.(2024)Xia, Malladi, Gururangan, Arora, and
  Chen]{xia2024less}
Mengzhou Xia, Sadhika Malladi, Suchin Gururangan, Sanjeev Arora, and Danqi
  Chen.
\newblock Less: Selecting influential data for targeted instruction tuning.
\newblock \emph{arXiv preprint arXiv:2402.04333}, 2024.

\bibitem[Xia et~al.(2023)Xia, {Jiale Liu}, {Jun Yu}, {Xu Shen}, {Bo Han}, and
  {Tongliang Liu}]{Moderate}
Xiaobo Xia, {Jiale Liu}, {Jun Yu}, {Xu Shen}, {Bo Han}, and {Tongliang Liu}.
\newblock Moderate coreset: A universal method of data selection for real-world
  data-efficient deep learning.
\newblock In \emph{International Conference on Learning Representations}, 2023.

\bibitem[Yang et~al.(2023)Yang, {Zeke Xie}, {Hanyu Peng}, {Min Xu}, {Mingming
  Sun}, and {Ping Li}]{OPT}
Shuo Yang, {Zeke Xie}, {Hanyu Peng}, {Min Xu}, {Mingming Sun}, and {Ping Li}.
\newblock Dataset pruning: Reducing training data by examining generalization
  influence.
\newblock In \emph{International Conference on Learning Representations}, 2023.

\bibitem[Yang et~al.(2024)Yang, Cao, Guo, Zhang, Luo, Zhang, and
  Nie]{yang24icmlccsimporve}
Shuo Yang, Zhe Cao, Sheng Guo, Ruiheng Zhang, Ping Luo, Shengping Zhang, and
  Liqiang Nie.
\newblock Mind the boundary: Coreset selection via reconstructing the decision
  boundary.
\newblock In \emph{Proceedings of the 41st International Conference on Machine
  Learning}, volume 235, pp.\  55948--55960, 2024.

\bibitem[Yu et~al.(2020)Yu, Chan, You, Song, and Ma]{yu2020learning}
Yaodong Yu, Kwan Ho~Ryan Chan, Chong You, Chaobing Song, and Yi~Ma.
\newblock Learning diverse and discriminative representations via the principle
  of maximal coding rate reduction.
\newblock \emph{Advances in Neural Information Processing Systems},
  33:\penalty0 9422--9434, 2020.

\bibitem[Yu et~al.(2022)Yu, Khadivi, and Xu]{yu2022can}
Yu~Yu, Shahram Khadivi, and Jia Xu.
\newblock Can data diversity enhance learning generalization?
\newblock In \emph{Proceedings of the 29th international conference on
  computational linguistics}, pp.\  4933--4945, 2022.

\bibitem[Zheng et~al.(2022)Zheng, Liu, Lai, and Prakash]{CCS}
Haizhong Zheng, Rui Liu, Fan Lai, and Atul Prakash.
\newblock Coverage-centric coreset selection for high pruning rates.
\newblock 2022.

\end{thebibliography}
\bibliographystyle{main}

\newpage
\appendix

\section*{Broader Impact}
This paper presents a novel and effective data pruning algorithm to advance the deep learning area. There are some potential positive societal effects, such as helping people better understand the role of data to develop more robust deep learning systems and possibly even be used to reduce training and storage costs. Additionally, with the explosive growth of data, similar data-centric deep learning schemes can also be easily applied to inhumane surveillance or scenarios that violate data privacy and data copyright. Therefore, we believe that strict legislation is needed to constrain the occurrence of these.

\section*{Limitations} This paper proposes an efficient and high-performance coreset selection scheme. However, due to the limitations of experimental equipment, the maximum scale of the experiments in this paper only reaches a data scale of 12M. In the future, we will consider investing more in renting or purchasing more computing equipment and exploring the performance boundaries of InfoMax on a larger scale (at the billion level) dataset.

\begin{table*}[thtp]  
\caption{Comprehensive overview of the notational convention. \label{tb:notation}}
\begin{tabularx}{0.99\linewidth}{lX}
\toprule
Notation~~\quad~~~~~~~&Description\\
\midrule
$\mathbf{D}$        & The training set, and the size of the training set is $|\mathbf{D}|=N$.\\
$\mathbf{S} \subset \mathbf{D}$        & A subset from $\mathbf{D}$.\\
$\mathbf{S}^* \subset \mathbf{D}$        & The optimal coreset.\\
$p$        & The coreset budget and the target coreset size.\\
$z \in \mathbf{D}$        & A training data.\\
$I(z)$        & The information of sample $z$.\\
$I(\mathbf{S})$        & The set-level information of the set $\mathbf{S}$.\\
$I(z|...)$        & {The conditional information gain of sample z}.\\
$I(z_1, ..., z_n)$        & The set-level information of the set $[z_1, ..., z_n]$.\\
$I(z;s)$        & The mutual information between sample $z$ and $s$.\\
$\mathbf{I} \in \mathbb{R}^N$        & The information vector, where each item measures the informativeness of the corresponding sample, that is, $\mathbf{I}_z = I(z)$.\\
$\mathbf{K} \in \mathbb{R}^{N\times N}$        & The similarity matrix measures the similarity between samples.\\
$\mathbf{K}_{z,s}$        & {The similarity between sample $z$ and sample $s$}.\\
$k$        & The sparse rate of $\mathbf{K}$.\\
$d$        & The dataset partitioning rate.\\
$\alpha$        & The pairwise weight in the objective of InfoMax, see Eq.~(\ref{eq: quadratic}).\\
$\lambda$        & The hyper-parameter in the Graph-Cut Conditional Gain measurement \citep{Submodularity_3}, see Eq.~(\ref{eq: GCCG}).\\
$\mathbf{X} \in \{0,1\}^{N}$        & The binary variable in Eq~\ref{eq: quadratic} (before continuous relaxation).\\
$\mathbf{X}_z$        & {The variable item corresponding to the sample z}.\\
$\mathbf{X}^t \in \mathbb{R}^{N}_{+}$       & The iterant variable in the $t$-th iteration (after continuous relaxation).\\
$T$       & The maximum iteration number of the InfoMax solver.\\
\bottomrule 
\end{tabularx}
\end{table*}

\begin{algorithm}[thtp]
\caption{InfoMax Coreset Selection.}
\label{alg}
\begin{algorithmic}[1]
{
\STATE \vspace{0.3mm} \textbf{Input:} A dataset $\mathbf{D}$ with $N$ samples, the division coefficient $d$, the target coreset size budget $p$.  
\STATE \vspace{0.3mm} \textbf{Initialization:} Set the coreset as a null-set $\mathbf{S}^{*} = \emptyset$; Divide the dataset $\mathbf{D}$ into $d$ random subsets $[\mathbf{D}_1, ..., \mathbf{D}_d]$, where each subset is size of $N/d$. 
\STATE \vspace{0.3mm} \textbf{Initialization:} Uniformly initialize the initial guess $\mathbf{X}^0 = [\frac{1}{N}]^{N/d}$; \vspace{0.3mm}
\FOR{$i \in \{1, ..., d\}$} 
\STATE \vspace{0.3mm} \textcolor{teal}{// Objective construction: supervised or unsupervised mode.}  
\STATE \vspace{0.3mm} Calculate the intra-information information vector $\mathbf{I}$ and the similarity matrix $\mathbf{\mathbf{K}}$; 
\STATE \vspace{0.3mm} \textcolor{teal}{// Iteratively perform the InfoMax solver.}  
\FOR{$t \in \{0, ..., T-1\}$} 
\STATE \vspace{0.3mm} $\mathbf{X}^{t+1} = \text{Softmax}\Big(p\cdot\mathbf{I} - 2p\alpha\cdot \mathbf{\mathbf{K}} \mathbf{X}^{t} \Big)$;  
\ENDFOR  
\STATE \vspace{0.3mm} Append the samples corresponding to the top-k largest item in $\mathbf{X}^{T}$ into ${\mathbf{S}^{*}}$.
\ENDFOR 
\STATE \vspace{0.3mm} \textbf{Output:}  The InfoMax coreset ${\mathbf{S}^{*}}$. 
} 
\end{algorithmic}
\end{algorithm}

\newpage

\section{Experimental Settings}

\subsection{Image Classiication} 
We utilized Pytorch \cite{pytorch} to implement our method. Our experiments were conducted on a server equipped with 8 Tesla-V100 GPUs. Additionally, we maintained identical hyper-parameters and experimental settings for training before and after dataset pruning. To ensure fairness, we made sure that the number of iterations on the selected subset and the full set was the same by following \cite{CCS,d2pruning}. 
For CIFAR-100, we utilize the SGD optimizer with weight-decay set to 5e-4, a learning rate of 0.1, and a batch size of 128.
For TinyImageNet, we use the SGD optimizer with weight-decay set to 5e-4, a learning rate of 0.3, and a batch size of 64.
For ImageNet-1K, we use the SGD optimizer with weight-decay set to 1e-4, warmup for 5 epochs, a learning rate of 0.4, and a batch size of 256.
Regarding data augmentation, we solely adopt RandomResizedCrop and RandomHorizontalFlip for all experiments.

\subsection{Vision-language Pretraining} 

For coreset selection on the vision-language dataset CC12M \citep{CC12M}, all experiments are conducted on 2 servers with a totally of 16 NVIDIA V100 GPUs. The selected model is CLIP model \citep{CLIP}. We follow the settings provided in the original paper. 
Specifically, the CLIP model \cite{CLIP} is trained for 32 epochs with AdamW optimizer, weight decay 0.2, and a batch size of 2048. After 1 warmup epoch, the learning rate gradually decreases from 1e-4 following the cosine strategy.

\textit{Zero-shot ImageNet classification.} The CLIP model has two encoders, one for text and one for images. During the zero-shot classification process, text descriptions corresponding to the ImageNet classes are formulated. For example, if a class "dog" exists, a text description like "a picture of a dog" might be created. These text descriptions are encoded by the text encoder of CLIP to obtain text embeddings. At the same time, the images from the ImageNet dataset are encoded by the image encoder of CLIP to get image embeddings. Then, the similarity between each image embedding and all the text embeddings (representing different classes) is calculated. The image is classified into the class whose text embedding has the highest similarity to the image embedding. 

\textit{Linear Prob.} This is a technique used to evaluate and analyze the performance of a pre-trained model. For the CLIP model, linear probing involves adding a linear layer on top of the pre-trained CLIP model and then training only this linear layer while keeping the rest of the CLIP model's parameters fixed. 

\textit{Image-Text Retrieval.} This is a task where the goal is to find the most relevant text description for a given image or find the most relevant image for a given text description. 
Let us use the Image-to-Text Retrieval as an example. 
The image is encoded using the vision encoder. This results in an image embedding that represents the visual features of the image. Then, text documents are also encoded (using the text encoder) to obtain their respective text embeddings. The similarity between the image embedding and all the text embeddings is computed. The text with the highest similarity score is retrieved as the relevant description for the image.

\subsection{Instruction Tuning} 

The specific settings for LoRA fine-tuning are as follows: the Lora-rank is 64, bf-16 precision is used, the number of epochs is 4, the Lora-target-modules include q-proj, k-proj, v-proj, o-proj, the learning rate is $1e^{-05}$, the batch size is 8, the gradient accumulation steps is 16, and the AdamW optimizer is used. This experiments is conducted on a server with 8 A100 GPUs. As for the calculation of the LESS score, please refer to \citep{xia2024less} for details. Before computing the score, it performs LoRA fine-tuning on the LLM on the training dataset and then retains the randomly-projected LoRA gradient corresponding to each sample. The LESS score reflects how well the gradient of a training sample is consistent with the gradient of the target dataset we are concerned about. 

\section{Further Analysis}
\label{sec: Further Analysis}

\subsection{Generalization ability test} 

Here, we demonstrate the generalization ability of InfoMax. In the test for cross-model generalization ability, we observe that InfoMax's coreset exhibits superior generalization ability compared to other coresets.  Experimental results are reported in Table \ref{tab: Cross model generalization}. 
InfoMax achieves the best performance in the cross-model generalization ability test. Notably, when using unsupervised scores (SSP \citep{SSP}) and unsupervised features from DINO \citep{oquab2023dinov2}, InfoMax has better cross-model generalization ability compared to the supervised version. This also indicates the significance of using unsupervised base models to extract features during data selection.

\begin{table}[thtp]
\caption{\label{tab: Cross model generalization} Cross model generalization ability test, including the setting of ResNet to SENet, and the setting of  ResNet to EfficientNet-B0. The experiments are conducted on CIFAR-10 \citep{CIFAR}. 
InfoMax (unsupervised) uses the unsupervised scores (SSP \citep{SSP}) and unsupervised features from DINO \citep{oquab2023dinov2}, while all remaining methods are based on the EL2N score \citep{GraNd_EL2N} and the feature extractor trained on CIFAR-10.}
\setlength{\tabcolsep}{3.1pt}
\centering
\resizebox{0.65\linewidth}{!}{  
\begin{tabular}{l|ccc|ccc|cccccccccccc}
    \toprule[1pt]
    \textbf{Dataset} & \multicolumn{3}{c|}{\textbf{ResNet}} & \multicolumn{3}{c|}{\textbf{ResNet to SENet}} & \multicolumn{3}{c}{\textbf{ResNet to ENet-B0}} \\
    \cmidrule(lr){1-1} \cmidrule(lr){2-4} \cmidrule(lr){5-7} \cmidrule(lr){8-10}
    \textbf{Selection Rate} & \textbf{50\%} & \textbf{30\%} & \textbf{20\%} & \textbf{50\%} & \textbf{30\%} & \textbf{20\%} 
     & \textbf{50\%} & \textbf{30\%} & \textbf{20\%}\\
    \midrule[1pt]
    \textbf{Random} &93.4 &90.9 &88.0 &93.8 &92.9 &88.8 &90.0 &87.2 &84.6 \\
    \midrule[0.1pt]
    \textbf{Moderate} &92.6 &90.6 &87.3 &91.4 &88.3 &85.7 &87.1 &86.9 &82.2\\
    \textbf{CCS} &95.0 &93.0 &91.0 &92.7 &89.4 &88.1 &90.5 &86.7 &84.1\\
    \textbf{$\mathbf{D^2}$-Pruning} &93.3  &91.4  &87.1 &94.3  &91.4  &87.1 &93.3  &91.4  &87.1\\ 
    \midrule[0.1pt]
    \textbf{InfoMax} &94.1 &92.7 &89.1 &94.4 &93.3 &89.9 &92.5 &90.6 &87.8\\ 
    \textbf{InfoMax (Unsupervised)} &93.6 &92.2 &90.1 &94.4 &93.8 &90.5 &93.1 &90.8 &88.4\\ 
\bottomrule[1pt]
\end{tabular}
}
\end{table}

Furthermore, we assess the cross-setting generalization ability by varying the score (Forgetting \citep{forgetting} and Margin \citep{margin}) and the type of feature (unsupervised DINO features \citep{oquab2023dinov2}, unsupervised VQGAN features \citep{vqgan}). We observe that regardless of the configurations, InfoMax can achieve remarkably superior results compared to score-based approaches that only utilize score or schemes that conduct K-Median Coreset solely with features. This thoroughly demonstrates the cross-setting generalization ability of InfoMax. 

\begin{table}[thtp]
\caption{\label{tab: Cross setting generalization} Cross setting generalization ability test. The experiments are conducted on CIFAR-10 \citep{CIFAR}. The pruning ratio is 10\%. 
VQGAN features are obtained from the VQGAN model \citep{vqgan}. 
}
\setlength{\tabcolsep}{3pt}
\centering
\resizebox{0.54\linewidth}{!}{  
\begin{tabular}{l|ccccc|c|c|cc}
    \toprule
    Method & Accuracy\\
    \midrule
    Forgetting (Score-based) \citep{forgetting} & 46.3 \\
    Margin (Score-based) \citep{margin} & 34.3 \\
    Supervised feature (K-Median diversity-based) & 38.9 \\
    DINO feature (K-Median diversity-based) \citep{oquab2023dinov2} & 31.7 \\
    VQGAN feature (K-Median diversity-based) \citep{vqgan} & 28.2  \\
    InfoMax: forgetting + supervised feature & 89.4 \\
    InfoMax: margin + supervised feature &88.0 \\
    InfoMax: margin + DINO feature &85.4 \\
    InfoMax: margin + VQGAN feature &83.7 \\
\bottomrule
\end{tabular}
}
\end{table}

\subsection{Time cost test} 
\label{sec: time}

We study the time cost of InfoMax in Table \ref{tb: time}. There are two main stages for InfoMax, the first one is the objective construction stage, including inferencing on all data and calculating the similarity matrix $\mathbf{K}$ and calculating the sample-wise score $\mathbf{I}$, and the second stage is the optimizing stage, which iteratively running the solver defined in Eq.~(\ref{eq:update rule}). It is easy to find that although the computational efficiency of our method is slower than that of score-based schemes, it is still faster than $D^2$-Pruning \citep{d2pruning}. This is because InfoMax does not have a greedy selection process on a per-sample basis. 
For CC12M \citep{CC12M}, we divide the dataset into 10 subsets, each with a size of approximately 1.2 million. We run InfoMax in multiple threads on two server nodes with eight GPUs each to process these subsets. The overall time consumption is approximately 37 minutes. Note that this time is significantly longer than that for processing 1 million ImageNet data because multiple processes occupying the common disk I/O slow down the time efficiency. 
It is not difficult to observe that the main sources of the aforementioned time consumption are the first stage, including the inference process on all the data and the construction process of the similarity matrix $\mathbf{K}$ with the k-NN algorithm. The block of disk IO by multiple operations will further drag down the efficiency. Therefore, using a more advanced K-NN algorithm and a more advanced disk can significantly improve efficiency.  

\begin{table}[thtp]
\caption{\label{tb: time} A study on the time cost of InfoMax and two selected competitors, $D^2$-Pruning \citep{d2pruning} and score-based method (Entropy \citep{entropy}), for 1 million image data (the training set of ImageNet-1K \citep{imagenet}). 
}
\setlength{\tabcolsep}{3.1pt}
\centering
\resizebox{0.64\linewidth}{!}{  
\begin{tabular}{l|ccccc|c|c|cc}
    \toprule
    Name & Time cost (min)\\
    \midrule
    Stage-1 of InfoMax & 14.6 \\
    Stage-2 of InfoMax & 1.7 \\
    Overall of InfoMax & 16.1 \\
    \midrule
    $D^2$-Pruning \citep{d2pruning} &23 \citep{d2pruning}\\
   Score-based method (entropy \citep{entropy}) &4.1\\
\bottomrule
\end{tabular}
}
\end{table}

\section{Related Works}
\label{sec: related works}

\paragraph{Score-based methods.} The score-based techniques are the most popular data selection approaches. The EL2N score \citep{GraNd_EL2N} measures the data difficulty by computing the average of the $\ell_2$-norm error vector from a set of networks. 
GraNd \citep{GraNd_EL2N} measures the importance by calculating the expectation of the gradient norm. 
The Forgetting score \citep{forgetting} counts how many times a model changes its prediction from correct to incorrect for each example during the training process. 
Memorization \citep{memory} assigns a score to each example based on how much its presence or absence in the training set affects the model’s ability to predict it correctly.
Diverse ensembles \citep{ddd} gave a score to each sample based on how many models in a group misclassified it. 
\citep{SSP} proposed to use the distance between the sample and its corresponding cluster center as the importance score. 
Influence score \citep{tan2023data,xia2024less} measures the sample-wise leave-one-out retraining influence on the model's performance. The score-based approach often suffers from performance problems in application, especially when the pruning ratio is large.  
Some recent works have tried to solve this problem through various methods, for example, Moderate \citep{Moderate} suggested selecting data points with scores close to the score median. Note that Moderate \citep{Moderate} can use any selection criterion, such as EL2N score \citep{GraNd_EL2N}, as a basis. 
{Dyn-Unc \cite{Dyn_Unc} proposed an efficient uncertainty-based score with awareness of training dynamics. 
Some related works also use sample-wise scores \citep{CLIP,Sieve} to reflect the quality of multi-modality data.

\textbf{Diversity-based (Geometry-based) methods.} Traditionally, diversity-based coreset schemes are a very classic computer science problem \citep{lloyd1982least,tan2006cluster,coates2012learning,harpeled2004on,feldman2011unified,feldman2013turning,Dos}. These schemes aim to find a subset of data points that maximizes the diversity among the selected elements. \cite{k_center} applied greedy k-center to choose the coreset with good data coverage. \cite{yu2020learning,chan2022redunet} proposed to use the coding rate to model measure the diversity. \cite{yu2022can} formulates the problem of finding the most diverse subset into the problem of maximizing the dispersion or convex hull volume. In addition, some works proposed to prune data from the perspective of submodule theory \citep{Submodularity_1,Submodularity_2,Submodularity_3} and linear programming \citep{OPT} to ensure diversity. 

\textbf{Hybrid methods.} Solely using the diversity-based method alone can hardly perform satisfactorily because they do not consider the intra-sample information. Recently, some hybrid works have also attempted to introduce that score-based scheme into the diversity-driven pipelines to achieve better performance. 
CCS \citep{CCS} sets different sampling ratios for samples with different scores to enhance data coverage and balance both easy and hard samples. 
\cite{yang24icmlccsimporve} introduces reconstructing the classification boundary on the original dataset as a goal and brings it into the framework of CCS. 
BADGE \citep{AL5} is a diversity-based selection method in active learning that clusters the gradient embeddings of the current model using k-means and selects a subset from each cluster. 
$D^2$-Pruning \citep{d2pruning} views data pruning as a node selection problem based on Message-Passing on a graph, where the intra-sample information is utilized as the node values on the sample graph. One of the core steps of $D^2$-Pruning is the Inverse Message Passing operation, which iteratively performs a greedy sample selection step. In each iteration, it will select the sample with the highest score from all unselected candidates and then reduce the score of samples in the neighborhood to guarantee that highly redundant parts are not selected in the subsequent iterations. 
Among the above methods, $D^2$-Pruning is currently the most advanced solution in terms of performance. However, due to the heuristic or greedy nature of the algorithm, the result obtained by using is often suboptimal, check Figure \ref{fig: intro comparison performance} for details.

\section{Proof}
\label{sec: proof}

\subsection{Derivation of the InfoMax Solver}

Firstly, we restate the quadratic optimization problem of InfoMax: 
\begin{equation}
    \max_{\mathbf{X} \in \{0,1\}^N} \sum_{z\in \mathbf{D}} \mathbf{X}_z I(z) - \alpha\cdot \sum_{z, s\in \mathbf{D}} \mathbf{K}_{z, s} \mathbf{X}_z\mathbf{X}_s, ~~~~~~\text{s.t.}~~~~\sum_{z\in \mathbf{D}} \mathbf{X}_z = p, 
\end{equation}

Solving the quadratic problem defined in Eq~.\ref{eq: quadratic} directly can be extremely computationally burdensome and may even prove intractable since the budget size $p$ is generally on the order of tens of thousands or even millions \citep{crf2,CRF1}. Continuous relaxation simplifies the problem by relaxing some of the discrete constraints to continuous ones, reducing the complexity of the search space and making the problem more amenable to efficient optimization algorithms, such as the gradient-based methods \citep{crf2,CRF1}. 
Here, we also introduce a continuous relaxation of the problem, enabling the use of a gradient-based solver for more efficient optimization:
\begin{equation}
    \max_{\mathbf{X} \in \mathbb{R}_{+}^N} \sum_{z\in \mathbf{D}} \mathbf{X}_z I(z) - \alpha\cdot \sum_{z, s\in \mathbf{D}} \mathbf{K}_{z, s} \mathbf{X}_z\mathbf{X}_s, ~~~~~~\text{s.t.}~~~~\sum_{z\in \mathbf{D}} \mathbf{X}_z = p, 
    \label{eq: relaxed problem}
\end{equation}
{According to \citep{crf2,CRF1,crf3,tan2021proxy}, the continuous (complex and non-convex) problem in Eq.~(\ref{eq: relaxed problem}) could be optimized by solving a series of the following (convex) sub-problems:} 
\begin{equation}
    \mathbf{X}^{t+1} = \arg\min_{\mathbf{X} \in \mathbb{R}_{+}^N, \sum_{z\in \mathbf{D}} \mathbf{X}_z = p}  -\mathbf{X}^{\mathrm{T}} \Big( \mathbf{I} - 2 \alpha \mathbf{K} \mathbf{X}^{t} \Big) + \lambda h(\frac{\mathbf{X}}{p}) + \frac{1}{\beta} D(\mathbf{X}, \mathbf{X}^{t}), 
    \label{eq: sub problem}
\end{equation}
{where ${X}^{t}$ is the solution of the $t$-th sub-problem, $\Big( \mathbf{I} - 2 \alpha \mathbf{K} \mathbf{X}^{t} \Big)$ is the gradient of the objective in Eq.~(\ref{eq: relaxed problem}) at $\mathbf{X}^{t}$. We introduce the convex entropy function $h(\cdot)$ controlled by a factor $\lambda$ the regularization term. A large $\lambda$ value makes the problem easier to solve \citep{crf2,CRF1,crf3,tan2021proxy}, but it may deviate from the original problem. 
The proximal operator $D(\mathbf{X}, \mathbf{X}^{t})$ is an optional regularization term to prevent the solution difference between the two iterations is too large. 
Following tradition \citep{crf2,CRF1,crf3,tan2021proxy}, we use $D(\mathbf{X}, \mathbf{X}^{t}) = \frac{\mathbf{X}}{p} \log \frac{\mathbf{X}}{p} - \frac{\mathbf{X}}{p} \log \frac{\mathbf{X}^t}{p}$, which is the Kullback-Leibler divergence measure the discrepancy between $\frac{\mathbf{X}}{p}$ and $\frac{\mathbf{X}^t}{p}$. Note that due to the non-negativity of $\mathbf{X}$ and the property that the sum is a fixed value $p$, $\frac{\mathbf{X}}{p}$ has a probabilistic meaning. Each element of it represents the probability that each sample is selected. 
If we differentiate this convex sub-problem, the optimal solution is identified by solving the following equation:} 
\begin{equation}
    0 = -\Big(\mathbf{I} - 2 \alpha \mathbf{K} \mathbf{X}^{t} \Big) + \lambda \Big( -\frac{1}{p} \log\frac{\mathbf{X}^{*}}{p} - 1 \Big)  + \frac{1}{\beta} \Big( \frac{1}{p} \log (\frac{\mathbf{X}^{*}}{\mathbf{X}^{t}})  +   \frac{1}{p}  \mathbf{X}^{t} \Big),~~~~  \text{s.t.~~~} \mathbf{X}^{*} \in \mathbb{R}_{+}^N, \sum_{z\in \mathbf{D}} \mathbf{X}^{*}_z = p,
    \label{eq: sub problem solution}
\end{equation}
{This has an analytic solution to \citep{crf2,CRF1,crf3,tan2021proxy},} 
\begin{align}
    \hat{\mathbf{X}}^{*} &= \exp\Big(\frac{\beta p}{\lambda\beta-1} (\mathbf{I} - 2 \alpha \mathbf{K} \mathbf{X}^{t})  + \frac{1}{1-\lambda\beta}(\log \frac{\mathbf{X}^{t}}{p} - \mathbf{X}^{t}) + \frac{\lambda\beta p}{1-\lambda\beta}\Big), \label{eq: solution general 1}\\
    \mathbf{X}^{t+1} &= \mathbf{X}^{*} = \hat{\mathbf{X}}^{*} / (\sum_z \hat{\mathbf{X}}^{*}_z). 
\end{align}
{Since the third term in Eq.\ref{eq: solution general 1} is a constant, the solution could be formulated as the following form:}
\begin{equation}
    \mathbf{X}^{t+1} = \text{Softmax}\Big(  \frac{\beta p}{\lambda\beta-1} (\mathbf{I} - 2 \alpha \mathbf{K} \mathbf{X}^{t})  + \frac{1}{1-\lambda\beta}(\log \frac{\mathbf{X}^{t}}{p} - \mathbf{X}^{t})     \Big).
\end{equation}
{The mapping by the exponential function $\exp$ followed by the summation normalization is just equivalent to the Softmax operator which is widely used in deep learning. By just setting $\lambda = 1$ and setting $\beta \rightarrow \infty$, we have the most simplified iterative solution:}
\begin{equation}
    \mathbf{X}^{t+1} = \text{Softmax}\Big( p\cdot \mathbf{I} - 2 p\cdot \alpha \mathbf{K} \mathbf{X}^{t} \Big). 
    \label{eq: solution of sub problem}
\end{equation}
Note that the convergence rate of this solver is quite fast. In particular, the norm of the iterant difference $|\mathbf{X}^{t+1}-\mathbf{X}^{t}|$ converges at a rate of $\mathcal{O}(\frac{1}{T})$ according to \citep{crf3,tan2021proxy}.

\subsection{Prod: How does InfoMax find the most informative coreset?}
\label{sec: proof how info max}

We explain why InfoMax can find the most informative coreset from the perspective of information theory. First, we restate the information maximization formulation of data pruning defined in Eq.~(\ref{eq:true info}), 
$\mathbf{S}^{*} = \arg\max_{\mathbf{S} \subset D, |\mathbf{S}|=p} I(\mathbf{S})$, 
where $p$ is the size of the target coreset. The $I(\mathbf{S})$ measures the set-level information of the candidate subset $\mathbf{S}$, specifically, 
\begin{equation}
\begin{aligned}
I(\mathbf{S}) &= I(z_1) + I(z_2|z_1) +  ... + I(z_p |z_1, ..., z_{p-1})\\
&= I(z_1) + \sum_{2 \leq k}^{p} I(z_k |z_1, ..., z_{k-1}). 
\label{eq: information 2 proof}
\end{aligned}
\end{equation}
where $\{z_1, ..., z_p\} \in \mathbf{S}$ are all samples from the set. Note that this equation always holds regardless of the order of samples. 
The intra-sample information $I(z_1)$ of the sample $z_1$ could be instantiated by various sample-wise scores \citep{SSP,entropy,tan2023data,GraNd_EL2N}.  
Regarding the conditional gain term $I(z|Z)$, we refer to the recent progress in submodular information measures (SIM) \cite{Submodularity_1,Submodularity_2,Submodularity_3}, which present several instantiations for the submodular conditional gain. Here, we select the simplest yet effective instantiation, Graph Cut conditional gain (GCCG), 
\begin{equation}
\small
    I(z_k |z_1, ..., z_{k-1})=f(z_k) - 2\lambda \sum_{i<k} \mathbf{K}_{z_i,z_k}. \label{eq: GCCG}
\end{equation} 
It measures the dissimilarity between the sample $z_k$ and all conditional samples. Specifically, $\mathbf{K}_{z_i,z_k}$ measures the similarity between the sample $i$ and the sample $k$, $\lambda$ is an undetermined coefficient and is a hyperparameter in the system. The submodular function $f$ maps from a ground set to a real value, and we can simply set it with  $f(z)={I}(z)$. Hence, we have the following instantiation for the conditional gain term: $I(z_k |z_1, ..., z_{k-1})= I(z_k) - 2 \lambda \sum_{i<k} \mathbf{K}_{z_i,z_k}$. 
By substituting it into the Eq.~(\ref{eq: information 2 proof}), we can instantiate and reformulate
the set-level information as: 
\begin{equation} 
I(\mathbf{S}) = \sum_{z\in\mathbf{D}} I(z) - 2\lambda\sum_{z\neq s\in\mathbf{D}} \mathbf{K}_{z,s}. 
\label{eq: true info GCCG}
\end{equation} 
We introduce a set of binary variables $\mathbf{X} \in \{0, 1\}^N$ where $N=|\mathbf{D}|$ is the size of the whole training set. 
In the selection procedure, $\mathbf{X}_z=1$ indicates the sample $z$ was selected, otherwise it was pruned. 
By the problem definition in Eq.~(\ref{eq:true info}) and the set-level information formulation Eq.~(\ref{eq: information 2 proof}), we can transform the original information maximization problem in Eq.~(\ref{eq:true info}) into the following combinatorial optimization problem, 
\begin{equation}
\scriptsize
\begin{aligned}
    \max_{\mathbf{X}\in\{0,1\}^N,~|\mathbf{X}|=p}: ~~~~~&\sum_{\mathbf{S} \subset \mathbf{D}} \prod_{z\in\mathbf{S}}\mathbf{X}_{z} I(\mathbf{S}) \\
    &= \frac{1}{p!} \sum_{z_1\neq ... \neq z_p \in \mathbf{D}} \prod_{i\leq p}\mathbf{X}_{z_i} I([z_1, ..., z_n]) ~~\\
    &= \frac{1}{p!} \sum_{\mathbf{S} \subset \mathbf{D}} \prod_{z\in\mathbf{S}}\mathbf{X}_{z}\Big(\sum_{z\in\mathbf{S} } I(z) - 2\lambda\sum_{z\neq s \in\mathbf{S}} \mathbf{K}_{z, s} \Big), \\
    &= \frac{1}{p!} \sum_{z\neq z_1 ... \neq z_{p-1} \in \mathbf{D}} \mathbf{X}_{z} \mathbf{I}_{z} \prod_{1\leq i}^{p-1} \mathbf{X}_{z_i} - 2\lambda \frac{1}{p!} \sum_{z\neq s\neq z_1 ... \neq z_{p-2} \in \mathbf{D}} \mathbf{X}_{z} \mathbf{X}_s \mathbf{K}_{z,s} \prod_{1\leq i}^{p-2} \mathbf{X}_{z_i}\\
    &= \frac{1}{p!}\sum_{z\neq z_1 ... \neq z_{p-2} \in \mathbf{D}} \mathbf{X}_{z} \mathbf{I}_{z} \prod_{1\leq i}^{p-2} \mathbf{X}_{z_i} \Big( \sum_{z_{p-1}} \mathbf{X}_{z_{p-1}} - \sum_{j=1}^{p-2} \mathbf{X}_{z_j} \Big) \\
    &~~~~~~~~- 2\lambda \frac{1}{p!} \sum_{z\neq s\neq z_1 ... \neq z_{p-3} \in \mathbf{D}} \mathbf{X}_{z} \mathbf{X}_s\mathbf{K}_{z,s} \prod_{1\leq i}^{p-3} \mathbf{X}_{z_i} \Big( \sum_{z_{p-2}} \mathbf{X}_{z_{p-2}} - \sum_{j=1}^{p-3} \mathbf{X}_{z_j} \Big)\\
    &= \frac{1}{p!} \sum_{z\neq z_1 ... \neq z_{p-2} \in \mathbf{D}} \mathbf{X}_{z} \mathbf{I}_{z} \prod_{1\leq i}^{p-2} \mathbf{X}_{z_i} \Big( p - \sum_{j=1}^{p-2} \mathbf{X}_{z_j} \Big) \\
    &~~~~~~~~ - 2\lambda \frac{1}{p!} \sum_{z\neq s\neq z_1 ... \neq z_{p-2} \in \mathbf{D}} \mathbf{X}_{z} \mathbf{X}_s\mathbf{K}_{z,s} \prod_{1\leq i}^{p-3} \Big( p - \sum_{j=1}^{p-3} \mathbf{X}_{z_j} \Big)
\end{aligned}
\label{eq:target proof 1}
\end{equation}
Since $\mathbf{X}$ is binary, hence, 
\begin{equation}
\scriptsize
\begin{aligned}
    \max_{\mathbf{X}\in\{0,1\}^N,~|\mathbf{X}|=p}: ~~~~~&\sum_{\mathbf{S} \subset \mathbf{D}} \prod_{z\in\mathbf{S}}\mathbf{X}_{z} I(\mathbf{S}) \\
    &= \frac{1}{p!}\sum_{z\neq z_1 ... \neq z_{p-2} \in \mathbf{D}} \mathbf{X}_{z} \mathbf{I}_{z} \prod_{1\leq i}^{p-2} \mathbf{X}_{z_i} \Big( p - \sum_{j=1}^{p-2} \mathbf{X}_{z_j} \Big) \\
    &~~~~~~~~ - 2\lambda \frac{1}{p!} \sum_{z\neq s\neq z_1 ... \neq z_{p-2} \in \mathbf{D}} \mathbf{X}_{z} \mathbf{X}_s\mathbf{K}_{z,s} \prod_{1\leq i}^{p-3} \mathbf{X}_{z_i} \Big( p - \sum_{j=1}^{p-3} \mathbf{X}_{z_j} \Big)\\
    &= \frac{1}{p!} p \cdot \sum_{z\neq z_1 ... \neq z_{p-2} \in \mathbf{D}} \mathbf{X}_{z} \mathbf{I}_{z} \prod_{1\leq i}^{p-2} \mathbf{X}_{z_i} - \frac{1}{p!}(p-2) \cdot \sum_{z\neq z_1 ... \neq z_{p-2} \in \mathbf{D}} \mathbf{X}_{z} \mathbf{I}_{z} \prod_{1\leq i}^{p-2} \mathbf{X}_{z_i} \\
    &~ - 2p \cdot\lambda \frac{1}{p!} \sum_{z\neq s\neq z_1 ... \neq z_{p-2} \in \mathbf{D}} \mathbf{X}_{z} \mathbf{X}_s \mathbf{K}_{z,s} \prod_{1\leq i}^{p-3}\mathbf{X}_{z_i} + 2(p-3) \cdot\lambda \frac{1}{p!}\sum_{z\neq s\neq z_1 ... \neq z_{p-2} \in \mathbf{D}} \mathbf{X}_{z}  \mathbf{X}_s\mathbf{K}_{z,s} \prod_{1\leq i}^{p-3} \mathbf{X}_{z_i}\\
    &= 2 \frac{1}{p!} \cdot \sum_{z\neq z_1 ... \neq z_{p-2} \in \mathbf{D}} \mathbf{X}_{z} \mathbf{I}_{z} \prod_{1\leq i}^{p-2} \mathbf{X}_{z_i} - 6 \cdot\lambda \frac{1}{p!} \sum_{z\neq s\neq z_1 ... \neq z_{p-2} \in \mathbf{D}} \mathbf{X}_{z} \mathbf{X}_s \mathbf{K}_{z,s} \prod_{1\leq i}^{p-3}\mathbf{X}_{z_i}
\end{aligned}
\label{eq:target proof 2}
\end{equation}
By applying a similar reduction process to the rest variables, we have:
\begin{equation}
\scriptsize
\begin{aligned}
    \max_{\mathbf{X}\in\{0,1\}^N,~|\mathbf{X}|=p}: ~~~~~&\sum_{\mathbf{S} \subset \mathbf{D}} \prod_{z\in\mathbf{S}}\mathbf{X}_{z} I(\mathbf{S}) \\
    &= \frac{1}{p!}(p-1)! \cdot \sum_{z \in \mathbf{D}} \mathbf{X}_{z} \mathbf{I}_{z} - (p-2)! \cdot\lambda \sum_{z\neq s \in \mathbf{D}} \mathbf{X}_{z} \mathbf{X}_{s} \mathbf{K}_{z,s} \\
    &= \frac{(p-1)!}{p!} \Big( \sum_{z\in\mathbf{D}}  I(z) \cdot \mathbf{X}_z  - \frac{\lambda (p-2)! }{(p-1)!}\sum_{z\neq s \in \mathbf{D}} \mathbf{K}_{z, s} \cdot \mathbf{X}_z \mathbf{X}_s \Big),\\
\end{aligned}
\label{eq:target proof 3}
\end{equation}

where $(p)!$ is a factorial function of $p$, and $\lambda$ is a hyperparameter to be determined in Eq.~(\ref{eq: information 2}). And we have $\frac{(p-2)!}{(p-1)!} = \frac{1}{p-1}$. 
Hence, if we use $\alpha$ to indicate $\frac{\lambda (p-2)!}{(p-1)!}$, we obtain the quadratic programming problem as defined in InfoMax. Therefore, we have proved that under the premise of using the instantiation of Graph-cut conditional gain (GCCG) for the conditional gain term, solving the data pruning problem in Eq.~(\ref{eq:true info}) to find the most informative subset is equivalent to solving the quadratic problem defined in Eq.~(\ref{eq: quadratic}).

\end{document}